%% file: neurips_2025.tex
\documentclass{article}

\usepackage[final]{neurips_2025}

\usepackage[utf8]{inputenc} 
\usepackage[T1]{fontenc}    
\usepackage{hyperref}       
\usepackage{url}            
\usepackage{booktabs}       
\usepackage{amsfonts}       
\usepackage{nicefrac}       
\usepackage{microtype}      
\usepackage{xcolor}         

\usepackage{graphicx}
\usepackage{tabularx} 
\usepackage{booktabs} 
\usepackage{xcolor}
\usepackage{array}
\usepackage{amsmath}
\usepackage{cleveref}
\usepackage{verbatim}

\usepackage{rotating}

\usepackage{subfiles}

\title{HyPlaneHead: Rethinking Tri-plane-like Representations in Full-Head Image Synthesis}

\author{%
  Heyuan Li\footnotemark[1] \\
  SSE, CUHK (Shenzhen) \\
  \texttt{heyuanli@link.cuhk.edu.cn} \\
  \And
  Kenkun Liu \\
  SSE, CUHK (Shenzhen) \\
  \texttt{kenkunliu@link.cuhk.edu.cn} \\
  \And
  Lingteng Qiu\footnotemark[2] \\
  Tongyi Lab, Alibaba Inc.\\
  \texttt{qiulingteng.qlt@alibaba-inc.com} \\
  \And
  Qi Zuo \\
  Tongyi Lab, Alibaba Inc.\\
  \texttt{muyuan.zq@alibaba-inc.com} \\
  \And
  Keru Zheng \\
  SSE, CUHK (Shenzhen) \\
  \texttt{keruzheng@link.cuhk.edu.cn} \\
  \And
  Zilong Dong\\
  Tongyi Lab, Alibaba Inc.\\
  \texttt{list.dzl@alibaba-inc.com} \\
  \And
  Xiaoguang Han\footnotemark[3] \\
  SSE, CUHK (Shenzhen)  \quad FNii-Shenzhen \\
  Guangdong Provincial Key Laboratory of Future Networks of Intelligence \\
  \texttt{hanxiaoguang@cuhk.edu.cn } \\
}

\begin{document}

\maketitle
\begin{figure}[h]
{\includegraphics[width=1\linewidth]{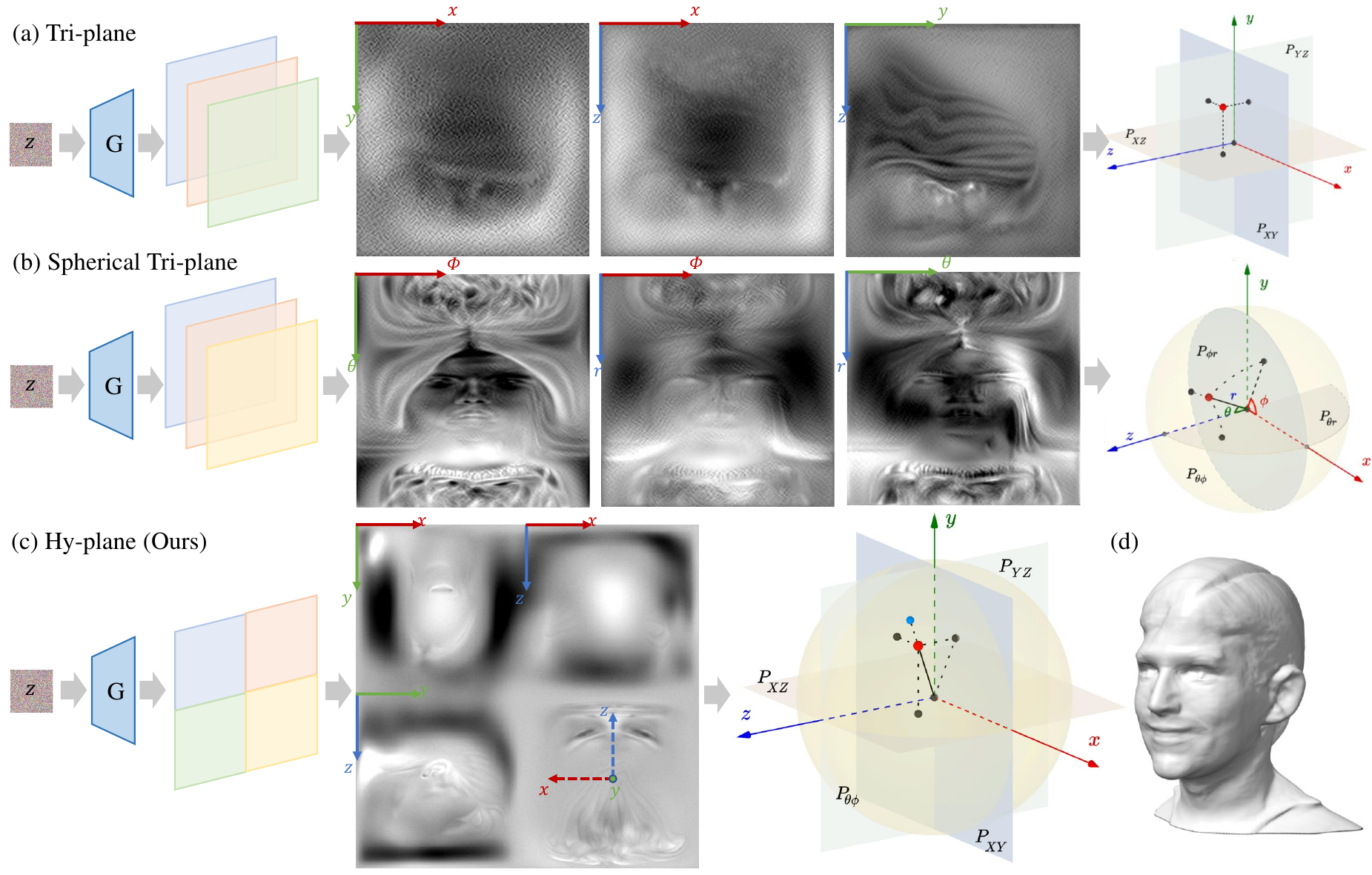}}
    \caption{
    {\small
    Figure (a, b, c) respectively illustrate the feature map visualizations and geometric structures of the tri-plane, spherical tri-plane, and our proposed hy-plane representation which integrates both planar and spherical planes.
    Figure (d) shows a head geometry model for defining the coordinate system.
    Note that in (a, b), the dominant planes ($P_{YZ}$ for tri-plane and $P_{\theta\phi}$ for spherical tri-plane) cause significant inter-channel feature penetration into the other two planes (best viewed when zoomed in), thereby limiting the model's expressiveness. In contrast, (c) resolves this issue entirely by employing a unify-split strategy, where all feature maps are generated within a single channel. As a result, each plane effectively learns its corresponding information without interference from other planes.
    }
    }
    \label{fig:teaser}
\end{figure}

\footnotetext[1]{Work done during an internship at Tongyi Lab, Alibaba Inc.}
\footnotetext[2]{Project lead.}
\footnotetext[3]{Corresponding author.}

\input{sec/0_abstract}    
\input{sec/1_intro}
\input{sec/2_related}

\input{sec/3_method}
\input{sec/4_experiments}
\input{sec/5_conclusion}
\input{sec/6_acknow}

\clearpage
{
\small
\bibliographystyle{ieeenat_fullname}
\bibliography{main}
}



\newpage
\section*{NeurIPS Paper Checklist}

\begin{enumerate}

\item {\bf Claims}
    \item[] Question: Do the main claims made in the abstract and introduction accurately reflect the paper's contributions and scope?
    \item[] Answer: \answerYes{} 
    \item[] Justification: Yes, the main claims made in the abstract and introduction accurately reflect the paper's contributions and scope.
    \item[] Guidelines:
    \begin{itemize}
        \item The answer NA means that the abstract and introduction do not include the claims made in the paper.
        \item The abstract and/or introduction should clearly state the claims made, including the contributions made in the paper and important assumptions and limitations. A No or NA answer to this question will not be perceived well by the reviewers. 
        \item The claims made should match theoretical and experimental results, and reflect how much the results can be expected to generalize to other settings. 
        \item It is fine to include aspirational goals as motivation as long as it is clear that these goals are not attained by the paper. 
    \end{itemize}

\item {\bf Limitations}
    \item[] Question: Does the paper discuss the limitations of the work performed by the authors?
    \item[] Answer: \answerYes{} 
    \item[] Justification: Although we do not include a dedicated section to discuss limitations due to page constraints, we briefly discuss the shortcomings of our method in \cref{sec:exp_quan}, such as the need for higher-resolution output feature maps and slightly slower inference speed compared to baseline methods.
    \item[] Guidelines:
    \begin{itemize}
        \item The answer NA means that the paper has no limitation while the answer No means that the paper has limitations, but those are not discussed in the paper. 
        \item The authors are encouraged to create a separate "Limitations" section in their paper.
        \item The paper should point out any strong assumptions and how robust the results are to violations of these assumptions (e.g., independence assumptions, noiseless settings, model well-specification, asymptotic approximations only holding locally). The authors should reflect on how these assumptions might be violated in practice and what the implications would be.
        \item The authors should reflect on the scope of the claims made, e.g., if the approach was only tested on a few datasets or with a few runs. In general, empirical results often depend on implicit assumptions, which should be articulated.
        \item The authors should reflect on the factors that influence the performance of the approach. For example, a facial recognition algorithm may perform poorly when image resolution is low or images are taken in low lighting. Or a speech-to-text system might not be used reliably to provide closed captions for online lectures because it fails to handle technical jargon.
        \item The authors should discuss the computational efficiency of the proposed algorithms and how they scale with dataset size.
        \item If applicable, the authors should discuss possible limitations of their approach to address problems of privacy and fairness.
        \item While the authors might fear that complete honesty about limitations might be used by reviewers as grounds for rejection, a worse outcome might be that reviewers discover limitations that aren't acknowledged in the paper. The authors should use their best judgment and recognize that individual actions in favor of transparency play an important role in developing norms that preserve the integrity of the community. Reviewers will be specifically instructed to not penalize honesty concerning limitations.
    \end{itemize}

\item {\bf Theory assumptions and proofs}
    \item[] Question: For each theoretical result, does the paper provide the full set of assumptions and a complete (and correct) proof?
    \item[] Answer: \answerNA{} 
    \item[] Justification: While this paper focuses primarily on practical implementation rather than theoretical analysis, we still provide the step-by-step formulas related to the geometric transformations in our novel near-equal-area warping strategy.
    \item[] Guidelines:
    \begin{itemize}
        \item The answer NA means that the paper does not include theoretical results. 
        \item All the theorems, formulas, and proofs in the paper should be numbered and cross-referenced.
        \item All assumptions should be clearly stated or referenced in the statement of any theorems.
        \item The proofs can either appear in the main paper or the supplemental material, but if they appear in the supplemental material, the authors are encouraged to provide a short proof sketch to provide intuition. 
        \item Inversely, any informal proof provided in the core of the paper should be complemented by formal proofs provided in appendix or supplemental material.
        \item Theorems and Lemmas that the proof relies upon should be properly referenced. 
    \end{itemize}

    \item {\bf Experimental result reproducibility}
    \item[] Question: Does the paper fully disclose all the information needed to reproduce the main experimental results of the paper to the extent that it affects the main claims and/or conclusions of the paper (regardless of whether the code and data are provided or not)?
    \item[] Answer: \answerYes{} 
    \item[] Justification: We provide a comprehensive description of the technical details of our method and experiments in the paper, which is sufficient for others to reproduce our results.
    \item[] Guidelines:
    \begin{itemize}
        \item The answer NA means that the paper does not include experiments.
        \item If the paper includes experiments, a No answer to this question will not be perceived well by the reviewers: Making the paper reproducible is important, regardless of whether the code and data are provided or not.
        \item If the contribution is a dataset and/or model, the authors should describe the steps taken to make their results reproducible or verifiable. 
        \item Depending on the contribution, reproducibility can be accomplished in various ways. For example, if the contribution is a novel architecture, describing the architecture fully might suffice, or if the contribution is a specific model and empirical evaluation, it may be necessary to either make it possible for others to replicate the model with the same dataset, or provide access to the model. In general. releasing code and data is often one good way to accomplish this, but reproducibility can also be provided via detailed instructions for how to replicate the results, access to a hosted model (e.g., in the case of a large language model), releasing of a model checkpoint, or other means that are appropriate to the research performed.
        \item While NeurIPS does not require releasing code, the conference does require all submissions to provide some reasonable avenue for reproducibility, which may depend on the nature of the contribution. For example
        \begin{enumerate}
            \item If the contribution is primarily a new algorithm, the paper should make it clear how to reproduce that algorithm.
            \item If the contribution is primarily a new model architecture, the paper should describe the architecture clearly and fully.
            \item If the contribution is a new model (e.g., a large language model), then there should either be a way to access this model for reproducing the results or a way to reproduce the model (e.g., with an open-source dataset or instructions for how to construct the dataset).
            \item We recognize that reproducibility may be tricky in some cases, in which case authors are welcome to describe the particular way they provide for reproducibility. In the case of closed-source models, it may be that access to the model is limited in some way (e.g., to registered users), but it should be possible for other researchers to have some path to reproducing or verifying the results.
        \end{enumerate}
    \end{itemize}

\item {\bf Open access to data and code}
    \item[] Question: Does the paper provide open access to the data and code, with sufficient instructions to faithfully reproduce the main experimental results, as described in supplemental material?
    \item[] Answer: \answerYes{} 
    \item[] Justification: We will open-source our code after the paper is accepted. In the supplementary material, we also submit key code related to the hy-plane representation and the near-equal-area warping strategy. The majority of the training data used comes from publicly available datasets, and similar results to those presented in the paper can be achieved using these public datasets.
    \item[] Guidelines:
    \begin{itemize}
        \item The answer NA means that paper does not include experiments requiring code.
        \item Please see the NeurIPS code and data submission guidelines (\url{https://nips.cc/public/guides/CodeSubmissionPolicy}) for more details.
        \item While we encourage the release of code and data, we understand that this might not be possible, so “No” is an acceptable answer. Papers cannot be rejected simply for not including code, unless this is central to the contribution (e.g., for a new open-source benchmark).
        \item The instructions should contain the exact command and environment needed to run to reproduce the results. See the NeurIPS code and data submission guidelines (\url{https://nips.cc/public/guides/CodeSubmissionPolicy}) for more details.
        \item The authors should provide instructions on data access and preparation, including how to access the raw data, preprocessed data, intermediate data, and generated data, etc.
        \item The authors should provide scripts to reproduce all experimental results for the new proposed method and baselines. If only a subset of experiments are reproducible, they should state which ones are omitted from the script and why.
        \item At submission time, to preserve anonymity, the authors should release anonymized versions (if applicable).
        \item Providing as much information as possible in supplemental material (appended to the paper) is recommended, but including URLs to data and code is permitted.
    \end{itemize}

\item {\bf Experimental setting/details}
    \item[] Question: Does the paper specify all the training and test details (e.g., data splits, hyperparameters, how they were chosen, type of optimizer, etc.) necessary to understand the results?
    \item[] Answer: \answerYes{} 
    \item[] Justification: The training and testing details are thoroughly described in the paper and supplementary material.
    \item[] Guidelines:
    \begin{itemize}
        \item The answer NA means that the paper does not include experiments.
        \item The experimental setting should be presented in the core of the paper to a level of detail that is necessary to appreciate the results and make sense of them.
        \item The full details can be provided either with the code, in appendix, or as supplemental material.
    \end{itemize}

\item {\bf Experiment statistical significance}
    \item[] Question: Does the paper report error bars suitably and correctly defined or other appropriate information about the statistical significance of the experiments?
    \item[] Answer: \answerYes{} 
    \item[] Justification: Yes, our paper focuses on addressing the limitations of the tri-plane representation and demonstrates improvements in both generation quality and artifact reduction. For artifact reduction, we provide visual comparisons to illustrate the qualitative improvement. In terms of generation quality, we use FID as a quantitative evaluation metric and report comparisons with baseline methods as well as ablation study results.
    \item[] Guidelines:
    \begin{itemize}
        \item The answer NA means that the paper does not include experiments.
        \item The authors should answer "Yes" if the results are accompanied by error bars, confidence intervals, or statistical significance tests, at least for the experiments that support the main claims of the paper.
        \item The factors of variability that the error bars are capturing should be clearly stated (for example, train/test split, initialization, random drawing of some parameter, or overall run with given experimental conditions).
        \item The method for calculating the error bars should be explained (closed form formula, call to a library function, bootstrap, etc.)
        \item The assumptions made should be given (e.g., Normally distributed errors).
        \item It should be clear whether the error bar is the standard deviation or the standard error of the mean.
        \item It is OK to report 1-sigma error bars, but one should state it. The authors should preferably report a 2-sigma error bar than state that they have a 96\% CI, if the hypothesis of Normality of errors is not verified.
        \item For asymmetric distributions, the authors should be careful not to show in tables or figures symmetric error bars that would yield results that are out of range (e.g. negative error rates).
        \item If error bars are reported in tables or plots, The authors should explain in the text how they were calculated and reference the corresponding figures or tables in the text.
    \end{itemize}

\item {\bf Experiments compute resources}
    \item[] Question: For each experiment, does the paper provide sufficient information on the computer resources (type of compute workers, memory, time of execution) needed to reproduce the experiments?
    \item[] Answer: \answerYes{} 
    \item[] Justification: Yes, we report the GPU model and number used in our experiments.
    \item[] Guidelines:
    \begin{itemize}
        \item The answer NA means that the paper does not include experiments.
        \item The paper should indicate the type of compute workers CPU or GPU, internal cluster, or cloud provider, including relevant memory and storage.
        \item The paper should provide the amount of compute required for each of the individual experimental runs as well as estimate the total compute. 
        \item The paper should disclose whether the full research project required more compute than the experiments reported in the paper (e.g., preliminary or failed experiments that didn't make it into the paper). 
    \end{itemize}
    
\item {\bf Code of ethics}
    \item[] Question: Does the research conducted in the paper conform, in every respect, with the NeurIPS Code of Ethics \url{https://neurips.cc/public/EthicsGuidelines}?
    \item[] Answer: \answerYes{} 
    \item[] Justification: Yes, our research has been conducted in full compliance with the NeurIPS Code of Ethics.
    \item[] Guidelines:
    \begin{itemize}
        \item The answer NA means that the authors have not reviewed the NeurIPS Code of Ethics.
        \item If the authors answer No, they should explain the special circumstances that require a deviation from the Code of Ethics.
        \item The authors should make sure to preserve anonymity (e.g., if there is a special consideration due to laws or regulations in their jurisdiction).
    \end{itemize}

\item {\bf Broader impacts}
    \item[] Question: Does the paper discuss both potential positive societal impacts and negative societal impacts of the work performed?
    \item[] Answer: \answerYes{} 
    \item[] Justification: Yes, we discuss both the potential positive and negative societal impacts of our work in the supplementary material.
    \item[] Guidelines:
    \begin{itemize}
        \item The answer NA means that there is no societal impact of the work performed.
        \item If the authors answer NA or No, they should explain why their work has no societal impact or why the paper does not address societal impact.
        \item Examples of negative societal impacts include potential malicious or unintended uses (e.g., disinformation, generating fake profiles, surveillance), fairness considerations (e.g., deployment of technologies that could make decisions that unfairly impact specific groups), privacy considerations, and security considerations.
        \item The conference expects that many papers will be foundational research and not tied to particular applications, let alone deployments. However, if there is a direct path to any negative applications, the authors should point it out. For example, it is legitimate to point out that an improvement in the quality of generative models could be used to generate deepfakes for disinformation. On the other hand, it is not needed to point out that a generic algorithm for optimizing neural networks could enable people to train models that generate Deepfakes faster.
        \item The authors should consider possible harms that could arise when the technology is being used as intended and functioning correctly, harms that could arise when the technology is being used as intended but gives incorrect results, and harms following from (intentional or unintentional) misuse of the technology.
        \item If there are negative societal impacts, the authors could also discuss possible mitigation strategies (e.g., gated release of models, providing defenses in addition to attacks, mechanisms for monitoring misuse, mechanisms to monitor how a system learns from feedback over time, improving the efficiency and accessibility of ML).
    \end{itemize}
    
\item {\bf Safeguards}
    \item[] Question: Does the paper describe safeguards that have been put in place for responsible release of data or models that have a high risk for misuse (e.g., pretrained language models, image generators, or scraped datasets)?
    \item[] Answer: \answerYes{} 
    \item[] Justification: Yes, we discuss safeguards in the supplementary material.
    \item[] Guidelines:
    \begin{itemize}
        \item The answer NA means that the paper poses no such risks.
        \item Released models that have a high risk for misuse or dual-use should be released with necessary safeguards to allow for controlled use of the model, for example by requiring that users adhere to usage guidelines or restrictions to access the model or implementing safety filters. 
        \item Datasets that have been scraped from the Internet could pose safety risks. The authors should describe how they avoided releasing unsafe images.
        \item We recognize that providing effective safeguards is challenging, and many papers do not require this, but we encourage authors to take this into account and make a best faith effort.
    \end{itemize}

\item {\bf Licenses for existing assets}
    \item[] Question: Are the creators or original owners of assets (e.g., code, data, models), used in the paper, properly credited and are the license and terms of use explicitly mentioned and properly respected?
    \item[] Answer: \answerYes{} 
    \item[] Justification: Yes, in addition to our own code and data, all other external resources such as code and datasets used in this work are publicly available, and we have properly credited their creators, respecting the licenses and terms of use.
    \item[] Guidelines:
    \begin{itemize}
        \item The answer NA means that the paper does not use existing assets.
        \item The authors should cite the original paper that produced the code package or dataset.
        \item The authors should state which version of the asset is used and, if possible, include a URL.
        \item The name of the license (e.g., CC-BY 4.0) should be included for each asset.
        \item For scraped data from a particular source (e.g., website), the copyright and terms of service of that source should be provided.
        \item If assets are released, the license, copyright information, and terms of use in the package should be provided. For popular datasets, \url{paperswithcode.com/datasets} has curated licenses for some datasets. Their licensing guide can help determine the license of a dataset.
        \item For existing datasets that are re-packaged, both the original license and the license of the derived asset (if it has changed) should be provided.
        \item If this information is not available online, the authors are encouraged to reach out to the asset's creators.
    \end{itemize}

\item {\bf New assets}
    \item[] Question: Are new assets introduced in the paper well documented and is the documentation provided alongside the assets?
    \item[] Answer: \answerYes{} 
    \item[] Justification: Yes, all new assets introduced in this work are thoroughly documented in both the main paper and the supplementary material.
    \item[] Guidelines:
    \begin{itemize}
        \item The answer NA means that the paper does not release new assets.
        \item Researchers should communicate the details of the dataset/code/model as part of their submissions via structured templates. This includes details about training, license, limitations, etc. 
        \item The paper should discuss whether and how consent was obtained from people whose asset is used.
        \item At submission time, remember to anonymize your assets (if applicable). You can either create an anonymized URL or include an anonymized zip file.
    \end{itemize}

\item {\bf Crowdsourcing and research with human subjects}
    \item[] Question: For crowdsourcing experiments and research with human subjects, does the paper include the full text of instructions given to participants and screenshots, if applicable, as well as details about compensation (if any)? 
    \item[] Answer: \answerNA{} 
    \item[] Justification: This paper does not involve crowdsourcing nor research with human subjects.
    \item[] Guidelines:
    \begin{itemize}
        \item The answer NA means that the paper does not involve crowdsourcing nor research with human subjects.
        \item Including this information in the supplemental material is fine, but if the main contribution of the paper involves human subjects, then as much detail as possible should be included in the main paper. 
        \item According to the NeurIPS Code of Ethics, workers involved in data collection, curation, or other labor should be paid at least the minimum wage in the country of the data collector. 
    \end{itemize}

\item {\bf Institutional review board (IRB) approvals or equivalent for research with human subjects}
    \item[] Question: Does the paper describe potential risks incurred by study participants, whether such risks were disclosed to the subjects, and whether Institutional Review Board (IRB) approvals (or an equivalent approval/review based on the requirements of your country or institution) were obtained?
    \item[] Answer: \answerNA{} 
    \item[] Justification: This paper does not involve crowdsourcing nor research with human subjects.
    \item[] Guidelines:
    \begin{itemize}
        \item The answer NA means that the paper does not involve crowdsourcing nor research with human subjects.
        \item Depending on the country in which research is conducted, IRB approval (or equivalent) may be required for any human subjects research. If you obtained IRB approval, you should clearly state this in the paper. 
        \item We recognize that the procedures for this may vary significantly between institutions and locations, and we expect authors to adhere to the NeurIPS Code of Ethics and the guidelines for their institution. 
        \item For initial submissions, do not include any information that would break anonymity (if applicable), such as the institution conducting the review.
    \end{itemize}

\item {\bf Declaration of LLM usage}
    \item[] Question: Does the paper describe the usage of LLMs if it is an important, original, or non-standard component of the core methods in this research? Note that if the LLM is used only for writing, editing, or formatting purposes and does not impact the core methodology, scientific rigorousness, or originality of the research, declaration is not required.
    \item[] Answer: \answerNA{} 
    \item[] Justification: The core method development in this research does not involve LLMs as any important, original, or non-standard components.
    \item[] Guidelines:
    \begin{itemize}
        \item The answer NA means that the core method development in this research does not involve LLMs as any important, original, or non-standard components.
        \item Please refer to our LLM policy (\url{https://neurips.cc/Conferences/2025/LLM}) for what should or should not be described.
    \end{itemize}

\end{enumerate}


\clearpage
\input{sec/9_supp}

\end{document}

%% file: sec/0_abstract.tex
\vspace{-0.95em}
\begin{abstract}

Tri-plane-like representations have been widely adopted in 3D-aware GANs for head image synthesis and other 3D object/scene modeling tasks due to their efficiency. 
However, querying features via Cartesian coordinate projection often leads to feature entanglement, which results in mirroring artifacts. 
A recent work, SphereHead, attempted to address this issue by introducing spherical tri-planes based on a spherical coordinate system. 
While it successfully mitigates feature entanglement, SphereHead suffers from uneven mapping between the square feature maps and the spherical planes, leading to inefficient feature map utilization during rendering and difficulties in generating fine image details.
Moreover, both tri-plane and spherical tri-plane representations share a subtle yet persistent issue: feature penetration across convolutional channels can cause interference between planes, particularly when one plane dominates the others (see \cref{fig:teaser}). 
These challenges collectively prevent tri-plane-based methods from reaching their full potential. 
In this paper, we systematically analyze these problems for the first time and propose innovative solutions to address them. 
Specifically, we introduce a novel hybrid-plane (hy-plane for short) representation that combines the strengths of both planar and spherical planes while avoiding their respective drawbacks.
We further enhance the spherical plane by replacing the conventional theta-phi warping with a novel near-equal-area warping strategy, which maximizes the effective utilization of the square feature map.
In addition, our generator synthesizes a single-channel unified feature map instead of multiple feature maps in separate channels, thereby effectively eliminating feature penetration. 
With a series of technical improvements, our hy-plane representation enables our method, HyPlaneHead, to achieve state-of-the-art performance in full-head image synthesis.



\end{abstract}

%% file: sec/1_intro.tex
\section{Introduction}
\label{sec:intro}


Photorealistic full-head synthesis~\cite{zhuang2022mofanerf,park2021nerfies,canela2023instantavatar,he2024head360,doukas2021headgan} stands as a cornerstone technology for emerging applications in augmented/virtual reality avatars, immersive telepresence systems, and next-generation digital content creation. 
While modern 2D generative adversarial networks (GANs)~\cite{goodfellow2020generative,radford2015unsupervised,mao2017least,gulrajani2017improved,zhou2021cocosnet,kang2023scaling} achieve remarkable image quality in frontal face generation, their fundamental limitation in 3D scene modeling becomes apparent when synthesizing head images under arbitrary viewpoints. 

\begin{figure}[t!] 
    \centering
    \includegraphics
    [width=1.0\textwidth]
    {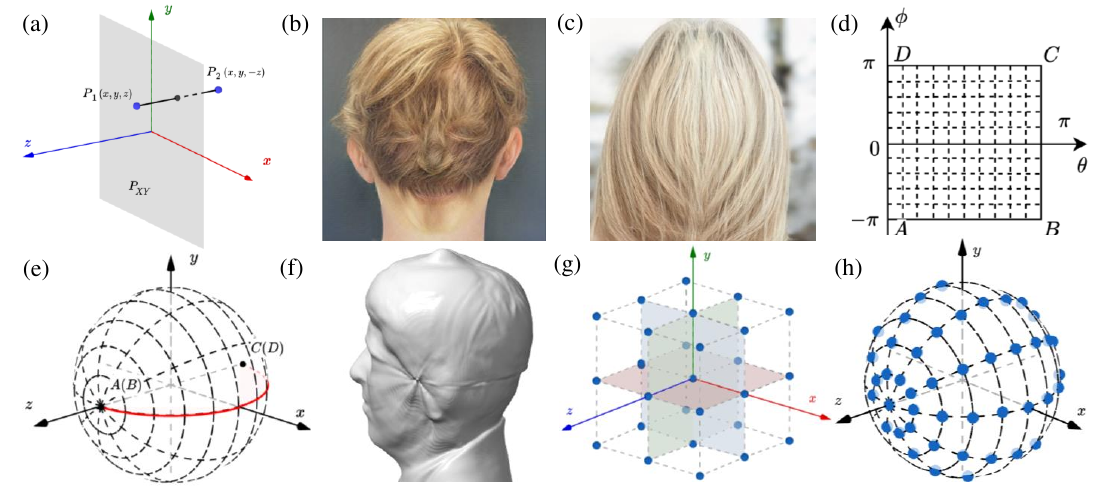} 
    \caption{
    {\small
    In the tri-plane representation, (a) the feature entanglement issue results in mirroring artifacts, where (b) the back of the head incorrectly exhibits front-face attributes, or (c) the hair's texture and shape display an unnaturally high degree of left-right symmetry.
    In the spherical tri-plane representation, (d, e) the non-equal-area warping caused by mapping a square to a sphere using $(\theta, \phi)$ coordinates introduces (f) artifacts in the seam and polar regions, as well as uneven spatial feature distribution after warping. 
    By contrast, (g) the tri-plane representation exhibits even spatial feature distribution, whereas (h) the spherical tri-plane representation shows uneven distribution, with features being overly dense in the polar regions and sparse near the equator.
    }
    }
    \label{fig:intro}
\end{figure}

Recent advancements in 3D-aware GANs~\cite{schwarz2020graf,deng2022gram,xue2022giraffe,nguyen2019hologan,chan2021pi,shi2021lifting,chan2022efficient,an2023panohead,li2024spherehead} have tackled this challenge by leveraging neural implicit representations, enabling view-consistent synthesis while maintaining photorealistic quality. 
Among these methods, the pioneering work EG3D~\cite{chan2022efficient} employs a tri-plane structure to represent human heads or other 3D objects. 
The tri-plane representation~\cite{gao2022get3d,shue20233d,zou2024triplane,wang2023rodin,hong2023lrm,gupta20233dgen,zuo2023dg3d,wu2024tpa3d,zuo2024high} efficiently captures symmetrical regions because two 3D points that are symmetric with respect to a feature plane will query the same feature on the plane via Cartesian coordinate projection. 
However, this inherent coupling of features becomes problematic in asymmetrical areas, leading to mirroring artifacts. 
As shown in \cref{fig:intro} (a, b), a typical example in full-head synthesis is that the back-view of the head shares the same features on the $P_{XY}$ plane as the front-view face, resulting in noticeable fake face artifacts on the back of the head. 
While PanoHead~\cite{an2023panohead} mitigates this issue by augmenting each plane with additional parallel planes, its tri-grid representation does not fundamentally resolve the problem, as it still inherits the same geometric and projection limitations of the Cartesian coordinate system.


A recent work, SphereHead~\cite{li2024spherehead}, creatively addresses the mirroring issue by introducing a spherical tri-plane representation that projects features in a spherical coordinate system. 
However, this approach introduces new challenges. 
First, it fails to leverage symmetry, which is prevalent in real-world objects. 
Second, the mapping from the square feature map to the spherical plane $P_{\theta \phi}$ involves a non-equal-area projection. 
Specifically, as illustrated in \cref{fig:intro} (c-f), after this mapping, features are sparsest near the equator and densest at the poles. 
This uneven distribution results in inefficient feature map utilization when rendering 2D images, reducing the model's ability to capture fine details. 
Moreover, referring to \cref{fig:intro} (d), a single spherical tri-plane can produce artifacts in the seam region due to the numerical discontinuity of $P_{\theta \phi}$ at $\phi=-\pi$ and $\phi=\pi$ in the spherical coordinate system. 
Although SphereHead mitigates this issue by incorporating an additional orthogonal spherical tri-plane, this solution complicates the model and introduces parameter redundancy. 
Worse still, the least expressive equatorial region of one sphere are used to cover the most expressive polar regions of the other, further diminishing the overall expressiveness of the representation.

Besides, we are the first to observe that a subtle yet persistent issue exists in both tri-plane and spherical tri-plane representations: feature penetration across convolutional channels can lead to interference between feature planes, particularly when one plane dominates the others.  
This issue arises because, unlike RGB images where channels are spatially aligned in 2D, each feature plane has a unique distribution, resulting in significantly different spatial meaning and values at the same uv position. 
In convolutional layers, however, all output channels at a given uv position are computed using the same input values. 
Ideally, the network is supposed to learn appropriate kernels to separate information for different planes. 
Yet, this is particularly challenging for 3D-aware GANs, as they are trained on 2D images without direct supervision on feature maps. 
Consequently, feature penetration often manifests visibly across planes, as shown in \cref{fig:teaser}.  
Although visible feature penetration gradually diminishes as training progresses, the issue itself remains difficult to fully resolve, subtly limiting the model's expressiveness and causing seemingly inexplicable artifacts.

In this paper, we introduce a simple yet effective \textbf{unify-split strategy} that generates a single-channel feature map and then splits it into multiple feature planes, instead of using different output channels to generate different feature planes. This approach completely eliminates the issue of feature penetration between channels.
Building on this, we propose a novel \textbf{hybrid-plane (hy-plane) representation} that integrates both planar and spherical feature planes, as illustrated in \cref{fig:teaser} (c). 
This design leverages the strengths of both tri-plane and spherical tri-plane representations while mitigating their respective limitations. Specifically, the hy-plane representation automatically learns symmetrical features using planar planes and captures anisotropic features through the spherical plane. 
This approach avoids mirroring artifacts while ensuring a uniformly high feature density throughout 3D rendering.
Furthermore, to optimize the mapping from the square feature map to the spherical plane, we employ Lambert azimuthal equal-area projection~\cite{ozturk2024lambert} combined with elliptical grid mapping~\cite{fong2019elliptificationrectangularimagery}. These techniques maximize the utilization of the square feature map and eliminate the seam artifacts.  
We also explore several variant models to further enhance performance. For instance, we increase the area proportion of the spherical plane to boost its expressive power and propose a dual-plane-dual-sphere variant to fully resolve polar artifacts. These innovations collectively contribute to the robustness and versatility of our hy-plane representation.


Building on the aforementioned technical advancements, the novel hy-plane representation enables our \textbf{HyPlaneHead} model to achieve state-of-the-art performance in full-head image synthesis, delivering high-quality results with significantly fewer artifacts compared to existing 3D-aware GAN methods~\cite{schwarz2020graf,deng2022gram,xue2022giraffe,nguyen2019hologan,chan2021pi,shi2021lifting,chan2022efficient,an2023panohead,li2024spherehead}.
In summary, our main contributions are as follows:



\begin{itemize}
    \item We conduct an in-depth analysis of the limitations inherent in tri-plane-like representations used in 3D-aware GANs. Based on this understanding, we introduce the \textit{hy-plane representation}, which combines the strengths of both planar and spherical planes while addressing their respective drawbacks.
    \item To achieve seamless integration of planar and spherical planes, we propose a series of technical innovations, including \textit{unify-split strategy}, a novel \textit{near-equal-area warping} method, \textit{area-biased splitting}, and exploration of \textit{alternative combination strategies}.
    \item Through comprehensive experiments, we validate the effectiveness of our proposed representation. Our \textit{HyPlaneHead} model achieves state-of-the-art performance for full-head image synthesis, demonstrating superior quality and reduced artifacts.
\end{itemize}

%% file: sec/2_related.tex
\section{Related Work}
\label{sec:related}


\noindent \textbf{3D Morphable Head Representations.} Traditional approaches for representing 3D faces with diverse shapes and appearances rely on 3D Morphable Models (3DMM)~\cite{blanz1999morphable, paysan20093d}, with FLAME~\cite{li2017learning} extending this framework to full head modeling. However, the coarse geometric details provided by 3DMMs have motivated numerous works to combine them with implicit neural representations, such as NeRF~\cite{canela2023instantavatar,zanfir2022phomoh,zheng2022imface,gafni2021dynamic,guo2021ad,park2021nerfies,wu2023ganhead,yenamandra2021i3dmm,zhang2023metahead,zhuang2022mofanerf}. While volume-based rendering techniques have significantly enhanced the capabilities of 3DMM-based models, their inherent topological constraints limit the expressiveness of implicit representations, particularly in capturing fine details like hair and wrinkles. Consequently, recent 3D-aware generative models have shifted towards directly synthesizing implicit neural fields of heads without relying on 3DMM priors.

\noindent \textbf{Generative Neural Head Representations.} Emerging neural head generative models~\cite{nguyen2020blockgan, schwarz2020graf,deng2022gram, chan2021pi, shi2021lifting, nguyen2019hologan, xue2022giraffe} adopt 3D-aware representations~\cite{mildenhall2021nerf}  which can be optimized by multi-view images through differentiable rendering. Though these implicit representations offer potential memory efficiency and structure complexity compared with traditional 3DMM-based representations~\cite{paysan20093d,blanz1999morphable,li2017learning}. Query-based feature sampling and fully connected mapping slow down the convergence process. To maintain representation complexity while accelerate the optimization process, EG3D~\cite{chan2022efficient} proposes tri-plane representation to explicitly store features on axis-aligned planes that are aggregated by a lightweight implicit feature decoder for efficient volume rendering. However, inherent coupling of features and mirroring issue are also brought with the efficiency. PanoHead~\cite{an2023panohead} propose to exploit extra in-the-wild data to supervise the back of head, thus can generate views in $360^o$ full head setting. Although it enriches the tri-plane’s representational capacity through adding more parallel feature planes, PanoHead can not thoroughly solve the mirroring issue from representation level. SphereHead~\cite{li2024spherehead}, through a shift in formulation from a Cartesian coordinate representation in cubic space to a spherical coordinate representation in spherical space, greatly eliminates the mirroring issue and avoids many artifacts. 
While SphereHead~\cite{li2024spherehead} addresses many limitations of prior work, it fails to preserve the simplicity of representing symmetric objects and introduces discontinuities along the seam between the two poles.
To this end, we propose a novel \textit{hy-plane representation} that effectively represents both symmetric and asymmetric regions, eliminates the mirroring issue, and avoids representation discontinuity.

%% file: sec/3_method.tex
\section{Method}
\label{sec:method}






\subsection{Hy-Plane Representation}
\label{sec:method.2}


As illustrated in \cref{fig:intro}, both tri-plane and spherical tri-plane have distinct strengths and limitations. 
The tri-plane representation benefits from uniform and dense spatial feature distribution, enabled by Cartesian coordinate projection. It efficiently renders high-resolution images from various angles and leverages symmetry effectively. However, it struggles with disentangling asymmetric features, leading to unwanted mirroring artifacts caused by feature entanglement. 
In contrast, the spherical tri-plane uses spherical coordinate projection to naturally distinguish directional features and learn anisotropic representations, avoiding feature entanglement. Yet, its non-uniform feature distribution reduces feature map utilization and complicates the rendering of high-resolution details.

Recognizing the complementary nature of these approaches, we propose \textit{hy-plane}, a novel hybrid representation combining planar and spherical planes. hy-plane uses planar components to capture symmetric features and spherical components to model anisotropic features. This design retains the efficiency and uniformity of the tri-plane while eliminating feature entanglement and mirroring artifacts.


\noindent \textbf{Hy-Plane (3+1)} 
The basic version of our representation consists of three planar planes plus one spherical plane, referred to as hy-plane (3+1). 
The three planar planes are arranged mutually orthogonally, with the positive z-axis aligned toward the human face, the positive y-axis pointing to the top of the head, and the positive x-axis directed toward the left ear. Features are queried using Cartesian coordinate projection. 
The spherical plane adopts a spherical coordinate system, with the polar axis aligned with the head's top direction. Notably, instead of directly querying the feature map using $(\theta, \phi)$ coordinates, we employ a novel \textit{near-equal-area warping} method to improve feature map utilization.



\begin{figure*}[t!] 
    \centering
    \includegraphics[width=\textwidth]{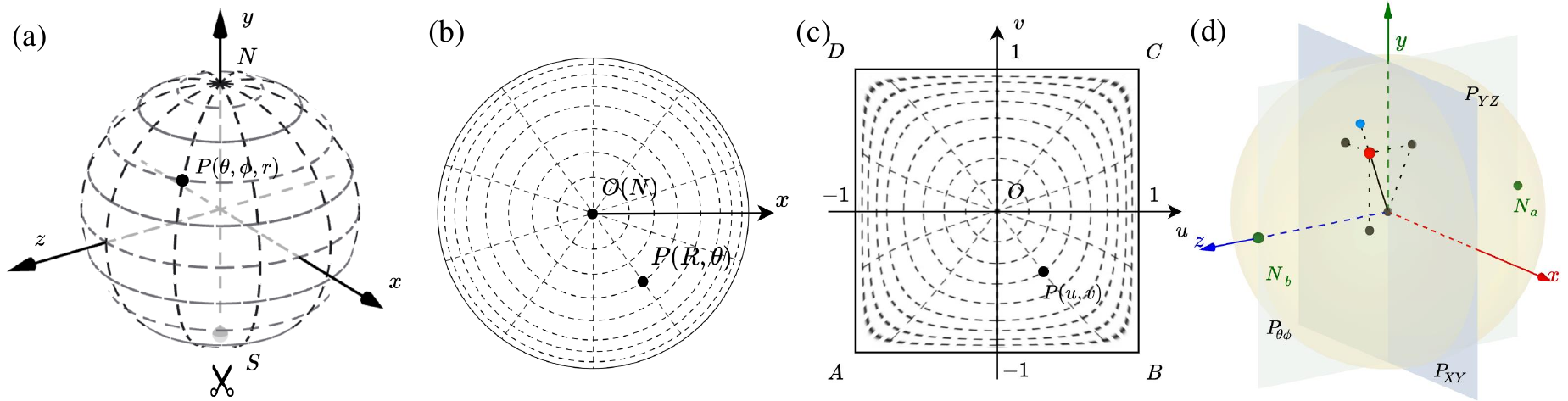} 
    \caption{
    {\small 
    (a) The Lambert azimuthal equal-area projection (LAEA) opens the South Pole, and maps the sphere to (b) a flat circular plane, with the North Pole at its center.
    (c) Elliptical grid mapping transforms the circular plane into a square. Conversely, the square can be inversely mapped back to a sphere with near-equal-area properties.
    (d) In the hy-plane (2+2) representation, two spheres coincide, with their North Poles facing in opposite directions. Please refer to the supplementary videos for a more comprehensive understanding.
    } 
    }
    \label{fig:warping}
\end{figure*}

\noindent \textbf{Near-Equal-Area Warping} 
Directly querying the feature map using $(\theta, \phi)$ coordinates, as in spherical tri-plane, is straightforward but introduces significant side effects. 
Geometrically (\cref{fig:intro}(d-f)), wrapping a square feature map $P_{\theta \phi}$ into a spherical plane causes numerical discontinuities at $\phi=-\pi$ and $\phi=\pi$, leading to artifacts in the seam region. Additionally, the edges $\theta=0$ and $\theta=\pi$ contract into polar points, converting fluctuations along these lines into high-frequency noise around the poles.
Furthermore, this warping is non-equal-area, unevenly distributing features from the square feature map onto the sphere. The equator region becomes feature-sparse, reducing expressive ability, while the poles become feature-dense, causing polar artifacts.
Although SphereHead \cite{li2024spherehead} addresses seam and polar artifacts by introducing an orthogonal dual-sphere setup, this approach doubles the number of feature planes but compromises the model's overall expressiveness, because each sphere uses its feature-sparse equator region to cover the other’s feature-dense polar regions, further limiting representation quality.


To address the challenging wrapping problem, we propose an elegant solution based on Lambert azimuthal equal-area projection (LAEA projection):
\begin{equation}
\begin{aligned}
(R,\Theta) & =\left(2\cos\frac{1}{2}\phi,-\theta\right),
\end{aligned}
\label{eq:laea projection}
\end{equation}
where $(\theta, \phi)$ denote colatitude and longitude in spherical coordinates on the spherical feature plane, and $(R,\Theta)$ denote radius and azimuth in polar coordinates on the flatten circle feature map.
As shown in \cref{fig:warping} (a, b), LAEA projection unfolds the spherical surface from the South Pole and flattens it into a circular plane centered on the North Pole. This method ensures equal-area transformation by adaptively adjusting latitudinal line density along the radius, achieving uniform feature distribution during warping. Additionally, it consolidates the seam and two poles of the spherical coordinate system into a single point, making them easier to handle. We align this point with the downward direction of the 3D head, which remains invisible in the rendering.

Next, we use elliptical grid mapping to transform the circle into a square, as illustrated in \cref{fig:warping} (b, c), which can be formulated as follows:

\begin{equation}
\begin{aligned}
(x,y) & =\left(R \cos \Theta,R \sin \Theta\right),
\end{aligned}
\label{eq:cartesian_to_polar}
\end{equation}

\begin{equation}
\left\{
\begin{aligned}
u &= \frac{1}{2}\sqrt{2 + x^2 - y^2 + 2\sqrt{2}x} - \frac{1}{2}\sqrt{2 + x^2 - y^2 - 2\sqrt{2}x}, \\
v &= \frac{1}{2}\sqrt{2 - x^2 + y^2 + 2\sqrt{2}y} - \frac{1}{2}\sqrt{2 - x^2 + y^2 - 2\sqrt{2}y}.
\end{aligned}
\right.
\label{eq:elliptical grid mapping}
\end{equation}


where $(x, y)$ are coordinates on the circle, and $(u, v)$ are coordinates on the square feature map.
This near-equal-area mapping minimizes severe deformation, preserving feature quality. 
When querying a 3D point’s feature on the spherical plane, we first convert its spherical coordinates $(\theta, \phi)$ to polar coordinates $(R,\Theta)$ on the wrapped circle using \cref{eq:laea projection}. We then transform these polar coordinates into 2D Cartesian coordinates $(x, y)$ via \cref{eq:cartesian_to_polar}, and finally map them to the corresponding$(u, v)$ location on the square feature map using \cref{eq:elliptical grid mapping}. This approach maximizes the utilization of the feature map while effectively eliminating seam artifacts.



\noindent \textbf{Hy-Plane (2+2)} 
While in human head modeling, we can hide the final pole (the South Pole) by orienting it downward, this approach may not be applicable to other scenes or objects where no such unimportant direction exists for hiding. This limits its broader applicability.
To address this, we introduce hy-plane (2+2), a variant consisting of two orthogonal planar planes and two spherical planes with opposing poles.
When querying a 3D point’s feature on the spherical planes, we compute features separately and combine them using a weighting function:


\begin{equation}
\left\{
\begin{aligned}
w_a &= (R_{a}^{\max} - R_a)^2, \\
w_b &= (R_{b}^{\max} - R_b)^2, \\
f_{\text{sph}} &= \frac{w_a f_a + w_b f_b}{w_a + w_b}
\end{aligned}
\right.
\label{eq:dualsphere_weight}
\end{equation}

Here, $R_a$ and $R_b$ represent the radii of the 3D point projected onto the two wrapped circles. 
$R_{a}^{max}$ and $R_{b}^{max}$ denote the radii of the circles.
The weights $w_a$ and $w_b$ are inversely proportional to these radii, peaking at the center ($R=0$) and decreasing toward the edges ($R=R^{max}$). This design optimizes the use of the feature map's flat central region while minimizing the impact of the distorted edge areas, effectively resolving artifacts at the poles.



The reason for reducing one planar plane while adding a spherical plane is to be compatible with the unify-split strategy, as will be explained in \cref{sec:method.3}. Notably, as demonstrated in~\cite{wang2023benchmarkinganalyzing3dawareimage}, two orthogonal planar planes can function nearly identically to three, since any two planes ($P_{XY}$, $P_{XZ}$, $P_{YZ}$) encompass all three coordinates $(x, y, z)$. 


\subsection{Unify-Split Strategy}
\label{sec:method.3}

\noindent \textbf{Feature Penetration across Channels} 
A key reason for the widespread adoption of tri-plane-like representations is their ability to represent 3D objects using 2D feature planes, whose data structure is similar to 2D images.
This allows researchers to directly leverage existing 2D image generation architectures for 3D-aware object synthesis. 
However, reusing these models directly, without adapting to the inherent differences between 2D RGB images and 3D-aware tri-plane-like representations, leads to a critical oversight.
In RGB images, the three channels represent different colors but share the same 2D spatial context. That is, the same uv position corresponds to the same spatial location, with only color variations across channels. This creates strong correlations during neural network training, enabling the network to first learn shared features layer by layer and then separate them into individual channels at the final output layer.

In contrast, in tri-plane-like representations, each plane encodes features from different spatial directions. Consequently, the same uv position on different planes corresponds to entirely distinct spatial meanings. Forcing convolutional networks to learn unrelated features at identical uv positions across planes increases learning complexity and causes feature entanglement between disparate spatial locations. 
This issue is particularly pronounced in 3D-aware GANs, where the model indirectly optimizes feature planes by learning from 2D images. The difficulty in disentangling information across feature planes leads to visible interference between output channels, resulting in unexpected artifacts in the generated images.



\begin{figure*}[t!] 
    \centering
    \includegraphics[width=\textwidth]{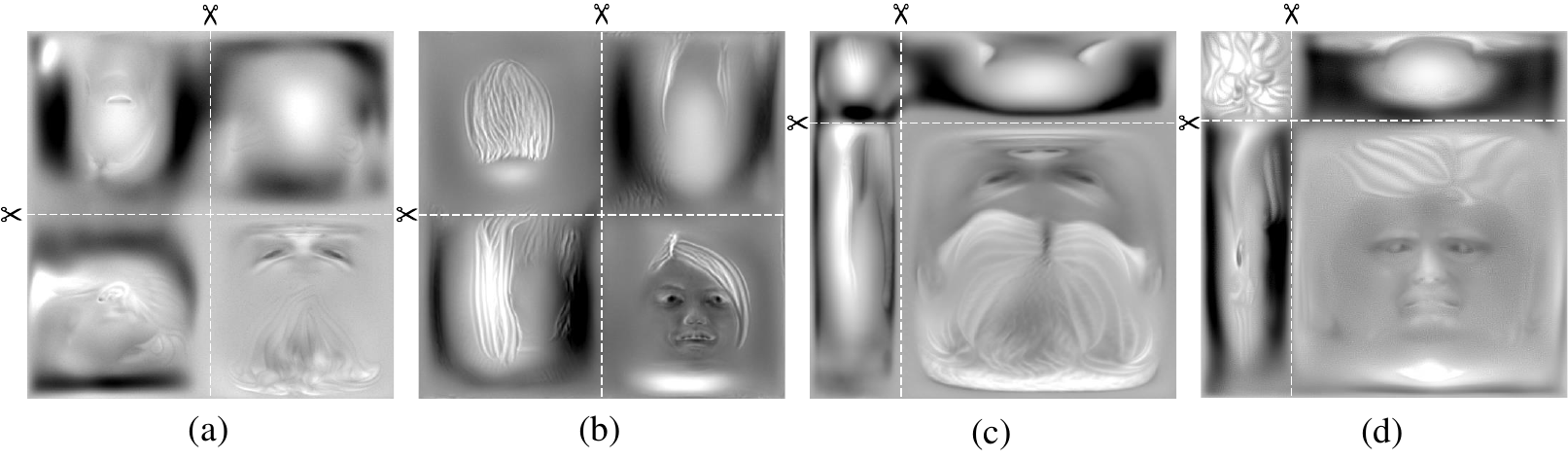} 
    \caption{
    {\small
    Unify-Split Strategy.  (a) Hy-plane (3+1) with evenly splitting; (b) Hy-plane (2+2) with evenly splitting; (c) Hy-plane (3+1) with area-biased splitting; (d) Hy-plane (2+2) with area-biased splitting.}
    }
    \label{fig:split}
\end{figure*}

\noindent \textbf{Evenly Splitting}  
Based on the aforementioned observation, we adopt a simple yet effective \textit{unify-split strategy} for synthesizing tri-plane-like representations. Instead of using separate channels for different feature planes, we allocate distinct regions on a large unified one-channel feature plane and then split it into parts corresponding to individual feature planes. This spatially disentangles features across planes in 2D space. \cref{fig:split} (a, b) illustrates the splitting process for hy-plane (3+1) and hy-plane (2+2), where all four planes are evenly divided into two-by-two configurations.

\noindent \textbf{Area-Biased Splitting}  
Additionally, we can refine the 2D splitting scheme to enhance specific capabilities of the hy-plane. For instance, in full-head synthesis, the ability to model anisotropic features is crucial for generating high-resolution back-head details. As shown in \cref{fig:split}(c, d), for hy-plane (3+1), we increase the area of the spherical plane and elongate the feature maps $P_{XY}$ and $P_{YX}$ along different axes. This maximizes their expressive power when combined.  
For hy-plane (2+2), we enlarge one primary spherical plane while shrinking the other. The larger plane remains a full sphere, while the smaller one forms a spherical cap, covering the problematic polar region of the larger sphere. 


\subsection{HyPlaneHead}
\label{sec:method.4}
We integrate our hy-plane representation into HyPlaneHead, a 3D-aware GAN pipeline akin to \cite{chan2022efficient,an2023panohead,li2024spherehead}.  
Given a sampled $z$ and conditioned camera parameter $c_{con}$, the generator $G$ produces a one-channel unified feature map, which is then split into individual feature planes of the hy-plane representation. Features are queried from each plane and volumetrically rendered using the viewing camera $c_{ren}$, enabling HyPlaneHead to generate a head image $I$ and mask $I^{m}$. As in \cite{chan2022efficient,an2023panohead,li2024spherehead}, the output passes through a super-resolution module to produce the high-resolution head image $I^{+}$. 
Following \cite{an2023panohead, li2024spherehead}, we also introduce a background generator to allow $ G $ to focus specifically on the head region.
In addition to the conventional 3D-aware GAN losses used in~\cite{chan2022efficient}, we further employ a view-image consistency loss, as proposed in~\cite{li2024spherehead}, to guide the discriminator to focus on the alignment between images and their corresponding viewpoints.


\begin{figure}[t!] 
    \centering
    \includegraphics[width=1.0\textwidth]{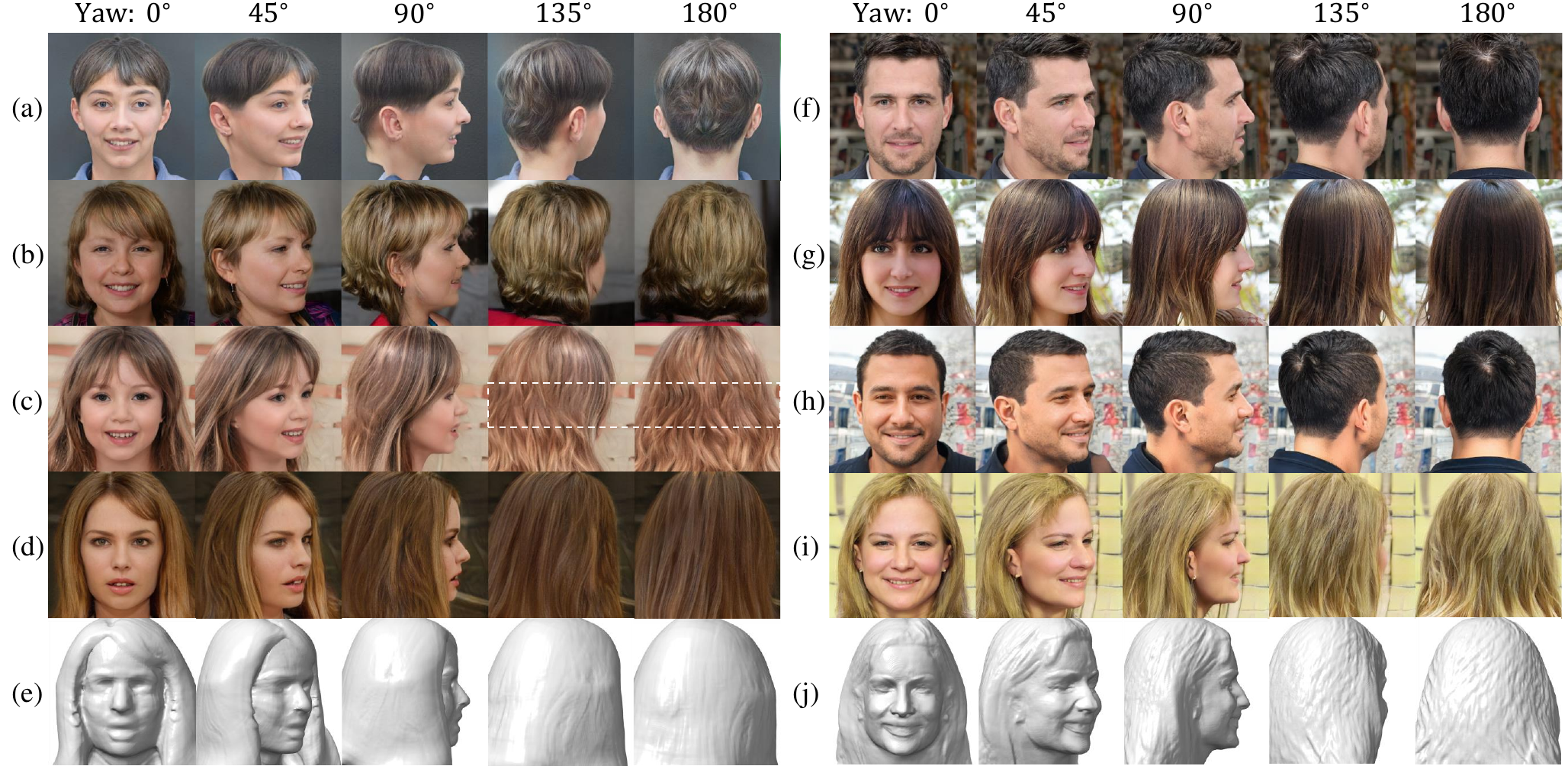} 
    \caption{
    {\small
    Qualitative comparison with state-of-the-art methods.
    (a) Tri-plane representation from \cite{chan2022efficient}.
    (b) Tri-grid representation from \cite{an2023panohead}.
    (c) Single spherical tri-plane representation from \cite{li2024spherehead}, where the white dashed box highlights a discontinuity in the hair region.
    (d, e) Dual spherical tri-plane representation from \cite{li2024spherehead}.
    (f–j) Our proposed Hy-plane representation.
    A closer inspection is recommended for finer details, and we refer readers to the supplementary material for higher-resolution visualizations.
    } 
    }
    \label{fig:exp_qual}
\end{figure}

%% file: sec/4_experiments.tex
\section{Experiments}
\label{sec:exp}

In this section, we conduct comprehensive qualitative and quantitative experiments on full-head image synthesis to demonstrate that our hy-plane representation is well-suited for rendering from any viewpoint. 
Our comparative analysis includes tri-plane, tri-grid, spherical tri-plane from EG3D, PanoHead, SphereHead respectively, and various hy-plane variants and settings for ablation study, all trained on our dataset and pipeline.
All experiments are trained on eight NVIDIA V100 GPUs with a batch size of 32. 
We follow PanoHead and SphereHead, using a training set that includes FFHQ~\cite{niemeyer2021campari}, CelebA~\cite{liu2018large}, LPFF~\cite{wu2023lpff}, WildHead~\cite{li2024spherehead}, K-Hairstyle~\cite{kim2021k}, and a 6K in-house dataset of large-pose head images processed with SphereHead's toolbox. 
All training images are 512×512 in resolution and augmented with horizontal flips. The entire training process spans 25 million images.
\subsection{Qualitative Comparison}
\label{sec:exp_quali}

We visualized synthesized samples and feature planes under varying configurations. For feature plane analysis, channel activations were averaged across feature-dimensions to visualize spatial texture patterns. 
As shown in \cref{fig:teaser}(a,b), tri-plane and spherical tri-plane models exhibit notable cross-plane interference: secondary feature maps show identical texture patterns from the dominating plane, alongside anomalous noise pattern. We suspect this phenomenon arises from inter-channel feature penetration, where competing planes disrupt each other's activations. Contrastingly, our unify-split strategy (\cref{fig:teaser}(c)) allows each plane to specialize in its directional features without cross-channel interference, thereby producing informative and clear feature maps.
From \cref{fig:teaser} (c) and \cref{fig:split}, our hy-plane representation effectively integrates planar and spherical planes, enabling seamless collaboration through a division of labor. The planar planes specialize in capturing symmetric features (e.g., $P_{YZ}$ learns left-right symmetric details such as side hair, ears, and shoulders), while the spherical plane excels at modeling anisotropic features (e.g., the frontal face and back-view hair). 

\cref{fig:exp_qual} compares our method with state-of-the-art full-head 3D-aware GANs.
(a) The tri-plane representation from \cite{chan2022efficient}, retrained on our dataset, generates mirrored faces at the back due to feature entanglement.
(b) PanoHead~\cite{an2023panohead} uses a tri-grid structure but still exhibits excessive symmetry in hairstyles.
(c) The single spherical tri-plane from SphereHead~\cite{li2024spherehead} produces full-head geometry but introduces seam and polar artifacts via ($\theta$, $\phi$) warping.
(d, e) Its dual spherical variant reduces artifacts but leads to over-smoothed textures and loss of detail.
(f–j) Our HyPlaneHead combines planar and spherical representations, achieving high-quality synthesis with rich texture and geometric fidelity, setting a new benchmark for full-head 3D-aware GANs.

\begin{table*}[t]\tiny
\centering
\caption{
{\small
Quantitative comparison on Full-head Image Synthesis task. We list FID, FID-random metrics with different representations. * denote that we output two spherical tri-planes simultaneously by a shared branch.
} 
}
\label{tab:full_head}
\resizebox{\textwidth}{!}{ 
\begin{tabular}{c c c c|c c}
\toprule
\textbf{No.}&\textbf{Representation} & \textbf{Unify-Split} & \textbf{Wrapping} & \textbf{FID} & \textbf{FID-random} \\
\hline
\hline
1& Tri-plane (EG3D \cite{chan2022efficient})      & -      & -      & 9.22      & 11.23   \\
2& Tri-plane      & evenly split      & -      & 8.86      & 11.52      \\ 
3 & Spherical Tri-plane (SphereHead \cite{li2024spherehead})      & -      & -      & 8.64      & 10.71      \\ 
4 & Spherical Tri-plane      & evenly split      & -      & 8.36      & 10.42     \\ 
5 & Dual Spherical Tri-plane      & -      & -      & 8.68      & 10.28     \\ 
6& Dual Spherical Tri-plane *   & -      & -      & 11.9      & 13.54     \\ 
7& Tri-grid (PanoHead \cite{an2023panohead})      & -      & -      & 8.77      & 10.66      \\ 
\hline
\hline
8& Tri-plane $512^{2}$     & -      & -      & 9.27      & 10.89     \\ 
9& Spherical Tri-plane $512^{2}$     & -      & -      & 8.82      & 10.47     \\ 
10& Tri-grid $512^{2}$     & -      & -      & 8.79      & 10.78     \\ 

\hline
\hline

11& Hy-plane (3+1)      & -      & -      & 8.54      & 10.66     \\ 
12& Hy-plane (3+1)      & evenly split      & -      & 8.31      & 10.18      \\ 
13& Hy-plane (3+1)      & evenly split      & yes      & 8.18      & 9.96      \\ 
14& Hy-plane (3+1)      & area-bias split      & yes      & \textbf{8.14}      & 9.88      \\ 
15& Hy-plane (2+2)      & evenly split      & yes      &  8.28     &    10.01   \\ 
16& Hy-plane (2+2)      & area-bias split      & yes      &  8.17    &   \textbf{9.84}   \\
\bottomrule
\end{tabular}
}
\end{table*}

\subsection{Quantitative Comparison and Ablation Study}
\label{sec:exp_quan}

To quantitatively evaluate the visual quality, fidelity, and diversity of the synthesized full-head images, we employed the Frechet Inception Distance (FID) metric~\cite{szegedy2016rethinking} on 50K real and synthetic samples. 
As noted in prior 3D-aware GANs based full-head image synthesis works~\cite{an2023panohead, li2024spherehead}, current 3D-aware GANs typically perform well under the conditioning camera pose during synthesis but degrade significantly at non-conditioned rendering angles.
To rigorously assess performance under arbitrary viewing angles, which is especially critical for full-head image synthesis, we introduced a new evaluation metric, FID-random, which decouples the conditioning pose from the rendering pose. Specifically, during generation, we first randomly sample a camera parameter $c_{con}$ from the dataset’s camera distribution to condition the tri-plane-like representation; subsequently, we render the head image using a different random camera parameter ${c_{ren}}$ (also sampled from the same distribution). The FID score is then calculated based on the images rendered under these random viewpoints, thereby providing an unbiased evaluation of the model’s robustness and generalization across all possible angles.


Comparing \cref{tab:full_head}(1) with \cref{tab:full_head}(11) demonstrates the advantages of augmenting the tri-plane with a spherical plane, consistent with our earlier visualizations. 
The effectiveness of our unify-split strategy is evidenced by the general reduction in FID and FID-random scores across \cref{tab:full_head}(1,2,3,4,11,12). Notably, while applying the unify-split strategy to the tri-plane reduces FID, it increases FID-random. This occurs because the strategy eliminates inter-channel feature penetration, allowing each plane to fully express its directional features. However, since the tri-plane does not separate directional features, the enhanced plane expression exacerbates mirroring artifacts on the backside, thereby worsening FID-random. In contrast, both the spherical tri-plane and hy-plane benefit from the separation of directional features provided by the spherical plane, enabling them to leverage the improved expressiveness unlocked by the unify-split strategy, resulting in reductions in both FID and FID-random.
\cref{tab:full_head}(3,4,5,6,9) reveal that directly outputting dual spherical tri-planes leads to significant interference between the two dominant theta-phi planes, yielding the highest FID scores. SphereHead mitigates this issue by introducing two small convolution-based branches, albeit at the cost of increased parameters. However, adopting the unify-split strategy achieves superior results without additional parameters, as demonstrated by the single spherical tri-plane's performance.
\cref{tab:full_head}(12,13) validate that wrapping improves performance by fully utilizing the square feature map. 
In \cref{tab:full_head}(14,16), we split a 512×512 feature map into four parts via area-bias splitting: 384×384, 384×128, 384×128, and 128×128, with the largest allocated to the spherical plane. Comparing these configurations with \cref{tab:full_head}(13,15) confirms the effectiveness of this partitioning scheme for full-head synthesis.
Finally, we tested the tri-plane, tri-grid, and spherical tri-plane with a feature map size of 512×512. \cref{tab:full_head}(8,9,10) show minimal impact from increasing the feature map size, ruling out model parameter scaling as a significant factor influencing our experimental outcomes.

\subsection{Single-view 3D-aware GAN Inversion}
\begin{figure}[t!] 
    \centering
    \includegraphics[width=1.0\textwidth]{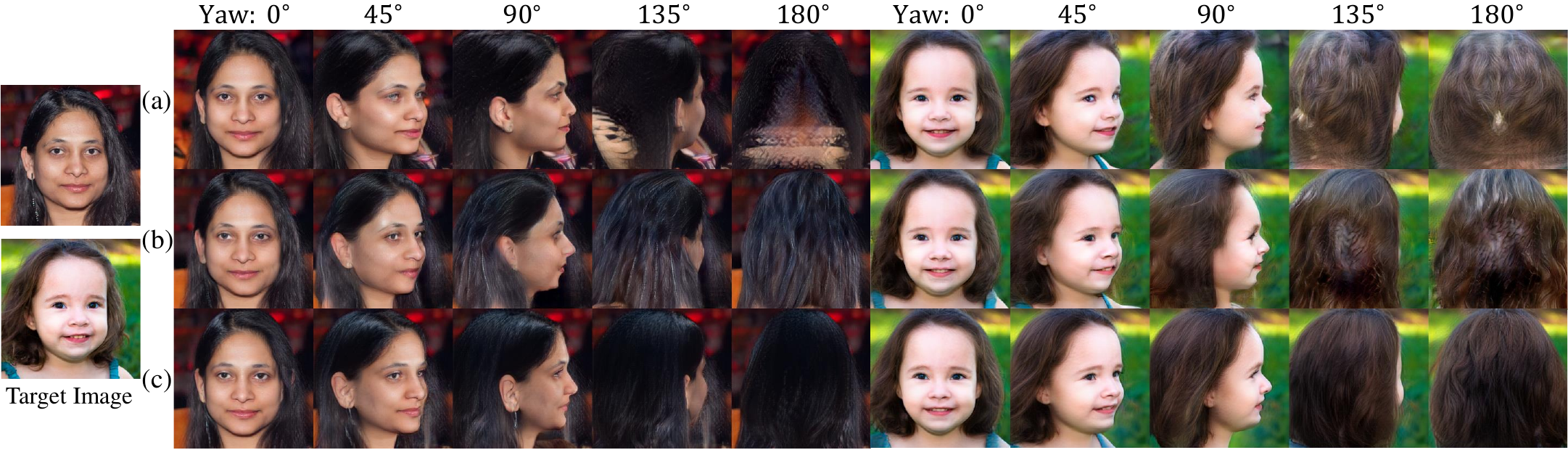} 
    \caption{
    {\small
    Single-view 3D-aware GAN inversion results. From top to bottom: 
    (a) PanoHead \cite{an2023panohead}, 
    (b) SphereHead \cite{li2024spherehead}, and 
    (c) our HyPlaneHead.
    } 
    }
    \label{fig:exp_inversion}
\end{figure}

We compare our method with PanoHead and SphereHead on 3D full-head reconstruction from a single-view image using Pivotal Tuning Inversion (PTI) \cite{roich2022pivotal}. As shown in \cref{fig:exp_inversion}, PanoHead consistently produces noticeable artifacts with strong left-right symmetry. SphereHead generates a plausible back-of-head region but yields blurry hair details, resulting in an overly coarse appearance. In contrast, our HyPlaneHead produces reasonable and high-quality renderings from all viewing angles.

%% file: sec/5_conclusion.tex
\section{Conclusion}


In this paper, we conduct an in-depth analysis of the limitations inherent in tri-plane-like representations used in 3D-aware GANs, particularly focusing on mirroring artifacts, uneven warping from the square feature map to the spherical plane, and feature penetration across channels. Based on these insights, we propose the hybrid-plane (hy-plane) representation, which combines the strengths of planar and spherical planes while mitigating their respective weaknesses. Our technical contributions include a unified planar-spherical representation, near-equal-area warping for seamless and efficient square-to-sphere mapping, and a unify-split strategy to eliminate feature penetration. These innovations enable HyPlaneHead to achieve state-of-the-art performance in full-head image synthesis, significantly reducing artifacts and enhancing rendering quality. 

%% file: sec/6_acknow.tex
\section{Acknowledgements}

The work was supported in part by NSFC with Grant No. 62293482, the Basic Research Project No. HZQB-KCZYZ-2021067 of Hetao Shenzhen-HK S\&T Cooperation Zone, Guangdong Provincial Outstanding Youth Fund(No. 2023B1515020055), by Shenzhen Science and Technology Program No. JCYJ20220530143604010, NSFC No.62172348, the Shenzhen Outstanding Talents Training Fund 202002, the Guangdong Research Projects No. 2017ZT07X152 and No. 2019CX01X104, the Guangdong Provincial Key Laboratory of Future Networks of Intelligence (Grant No. 2022B1212010001), and the Shenzhen Key Laboratory of Big Data and Artificial Intelligence (Grant No. SYSPG20241211173853027).

%% file: sec/9_supp.tex
\appendix

\begin{center}
  {\LARGE \textbf{Supplementary Material}}
\end{center}

\section{Overview}
This supplementary document provides additional materials to support the main paper \textit{"HyPlaneHead: Rethinking Tri-plane-like Representations in Full-Head Image Synthesis"}. 
We present high-resolution qualitative comparisons that highlight the performance of our method against existing approaches in \cref{sec:supp_B}.
A high-resolution qualitative comparison is provided in \cref{sec:supp_C}.
We also provide a detailed explanation of the Near-Equal-Area Warping technique and our proposed Hy-Plane (2+2) representation, which are central to achieving view-consistent and artifact-free full-head image synthesis in \cref{sec:supp_D}. 
Comprehensive measurements of parameter count, training and inference speed, and VRAM usage across various methods and HyPlaneHead configurations are reported in \cref{sec:supp_E}.
An in-depth discussion and comparison with related works, including OrthoPlanes, SYM3D, and other tri-plane-related algorithms, can be found in \cref{sec:supp_F}.
Additional qualitative results across a broader set of examples are included to further demonstrate the effectiveness of our model in \cref{sec:supp_G}.
We discuss the current limitations of our approach and potential directions for future work in \cref{sec:supp_H}. 
Finally, we include a section on Code of Ethics, where we address the ethical considerations and potential misuse of 3D head generation technologies in \cref{sec:supp_I}. 


\begin{figure}[h!]
    \centering
    \includegraphics[width=0.8\linewidth]{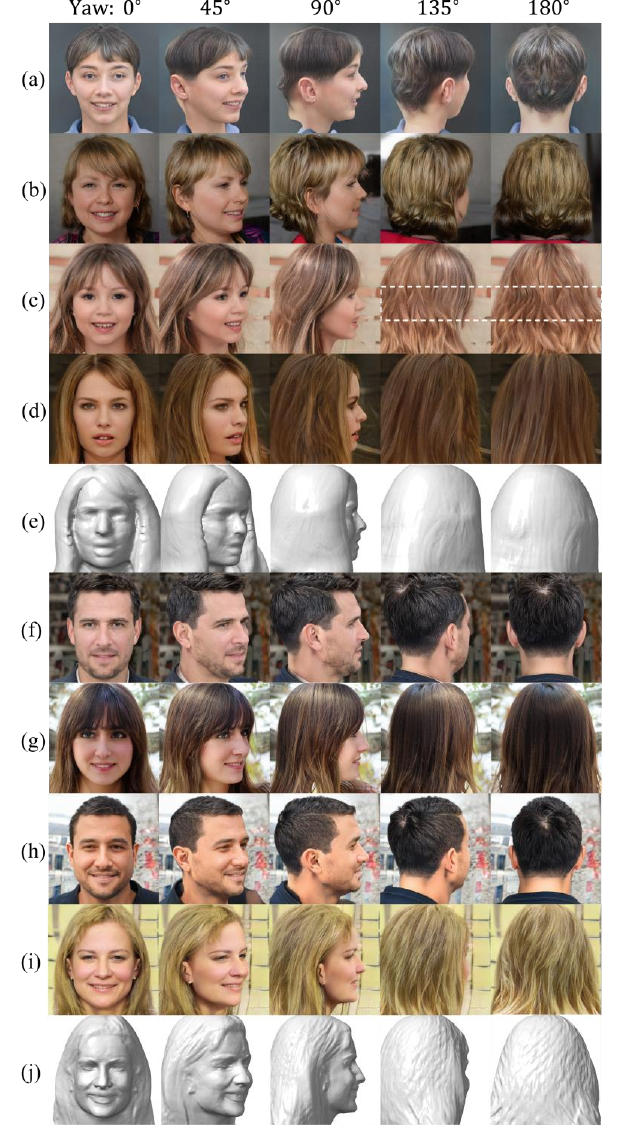}
    \caption{High-resolution qualitative comparison with state-of-the-art methods.}
    \label{fig:supp_fig1}
\end{figure}

\section{High-Resolution Qualitative Comparison (Main Paper Fig. 5)}
\label{sec:supp_B}
Due to the page limit of the paper, we were only able to include a low-resolution version of the qualitative comparison (Main Paper Fig. 5). To better demonstrate the advantages of our method in generating fine details, we provide a high-resolution version of Main Paper Fig. 5.

\cref{fig:supp_fig1} presents a high-resolution qualitative comparison with state-of-the-art methods. Each subfigure corresponds to the following representations: 
(a) Tri-plane representation from \cite{chan2022efficient}.
(b) Tri-grid representation from \cite{an2023panohead}.
(c) Single spherical tri-plane representation from \cite{li2024spherehead}, where the white dashed box highlights a discontinuity in the hair region caused by seam artifacts.
(d–e) Dual spherical tri-plane representation from \cite{li2024spherehead}.
(f–j) Our proposed Hy-plane representation.

While both the tri-plane and tri-grid representations (a and b) yield rich details in front-views, they suffer from inherent symmetry artifacts due to their Cartesian coordinate projections. Specifically, (a) exhibits clear mirroring face artifacts on the back of the head, reflecting front-view facial attributes. Similarly, (b) shows excessive left-right symmetry in the rear view.

The single spherical tri-plane (c) addresses the symmetry issue by introducing a spherical projection. However, it introduces seam artifacts due to the discontinuity in the $(\theta, \phi)$ warping at the boundary of the spherical feature map (as shown in the white-dashed box, where the hair texture is misaligned).

To mitigate these seams, the dual spherical tri-plane approach (d–e) introduces an additional orthogonal spherical tri-plane. While this effectively eliminates seam artifacts, it comes at the cost of increased parameter numbers. Moreover, when merging the two spherical tri-planes, the regions with the lowest feature density—i.e., the equatorial areas—are used to cover the high-density polar regions of the other plane. This results in reduced expressiveness for fine details such as hair textures and shape contours.

In contrast, our method employs the Hy-plane representation, which leverages the dense and even spatial feature distribution of the tri-plane, as well as its efficient representation of symmetric regions, to ensure high-fidelity detail reconstruction. At the same time, a spherical tri-plane is utilized to provide anisotropic representation for asymmetric areas, effectively eliminating mirroring artifacts. Furthermore, we introduce a novel near-equal-area sphere-to-square warping strategy that avoids seam artifacts without compromising detail preservation.

\section{Details of Near-Equal-Area Warping}
\label{sec:supp_C}
The Near-Equal-Area Warping method ensures that each region of the spherical input is mapped onto the planar representation with approximately equal surface area, add avoid excessive distortion and eliminating seam artifacts. This warping strategy is implemented in two steps: first, we use the Lambert Azimuthal Equal-Area Projection (LAEA projection) to flatten the spherical surface into a circular domain while preserving area; second, we apply Elliptical Grid Mapping to transform the circle into a square domain, enabling efficient utilization of the square-shaped feature map.

To better understand our proposed Near-Equal-Area Warping, we provide additional details and illustrative diagrams in this supplementary material.

\begin{figure}[t!]
    \centering
    \includegraphics[width=1.0\linewidth]{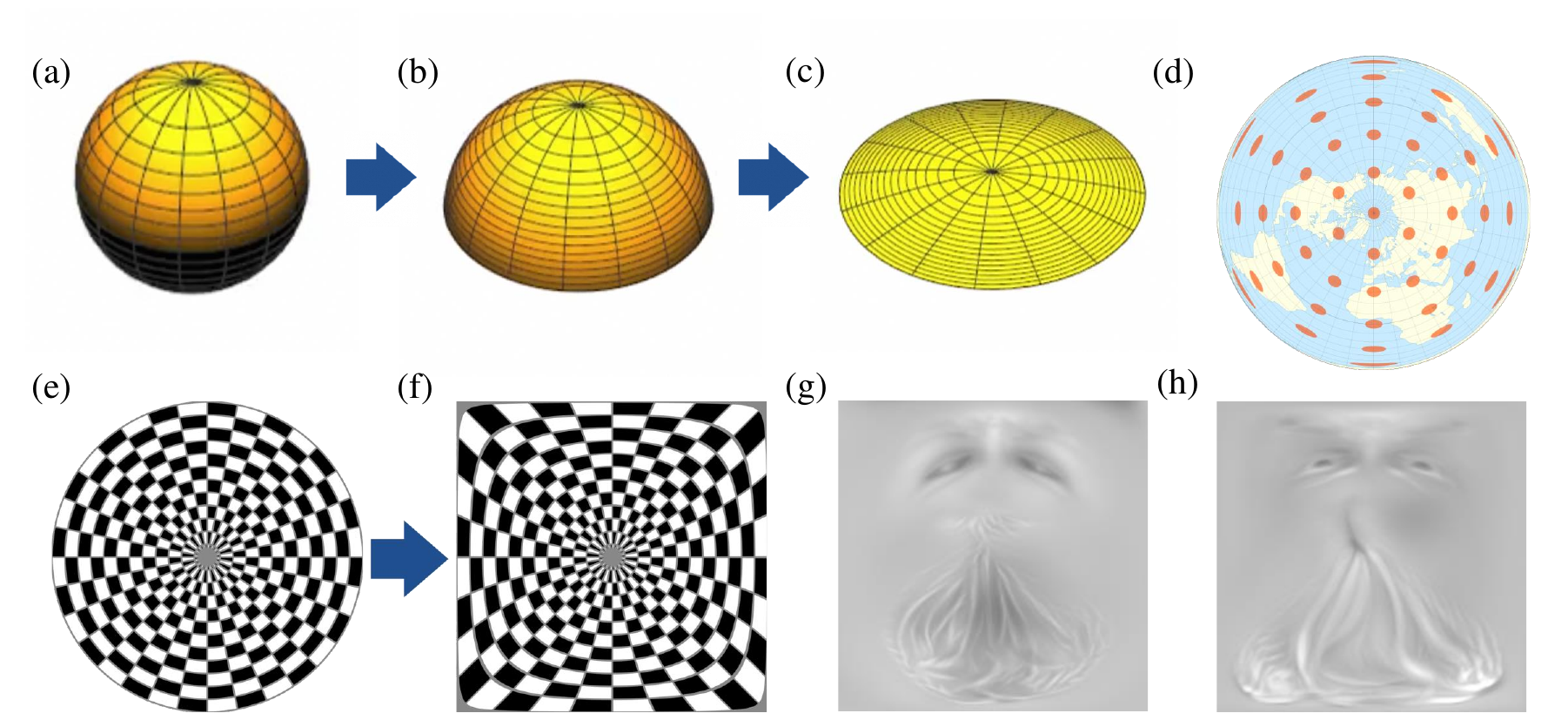}
    \caption{Illustrations of (a-d) the Lambert Azimuthal Equal-Area projection \protect\footnotemark[1] and (e-h) Elliptical Grid Mapping \protect\footnotemark[2].}
    \label{fig:supp_fig2}
\end{figure}

In the Lambert Azimuthal Equal-Area Projection, the south pole of the sphere is "opened" and then flattened into a circular domain centered at the north pole. During this unfolding process, the distances between latitude lines are adjusted such that the resulting circular projection maintains equal-area correspondence with the original spherical surface. A clearer understanding of this transformation can be gained from \cref{fig:supp_fig2}. 
\cref{fig:supp_fig2}(a–c) illustrates the dynamic process of the Lambert Azimuthal Equal-Area (LAEA) projection. \cref{fig:supp_fig2}(d) illustrates the Lambert Azimuthal Equal-Area (LAEA) projection using a world map example, demonstrating that it preserves area. Each orange circle represents a region of equal size on the original spherical surface.


Subsequently, we employ Elliptical Grid Mapping to convert the circular domain into a near-equal-area square grid. Among various methods for transforming a circle into a square, we choose Elliptical Grid Mapping due to its following advantageous properties: 
1. Approximate equal-area mapping: The variation in local area across the transformed plane is minimized. 
2. Smooth central region and minimal distortion at the boundaries: This preserves important structural details, especially near edges. 
3. Computationally simple and stable: It avoids division operations, which is crucial for maintaining gradient stability during training.
An intuitive illustration of this mapping is provided in \cref{fig:supp_fig2}. \cref{fig:supp_fig2}(e,f) show the deformation of Elliptical Grid Mapping under black-and-white stripe patterns, indicating that most regions experience minimal area distortion. 
\cref{fig:supp_fig2}(g,h) show the feature maps of the spherical plane before and after applying Elliptical Grid Mapping. Without this mapping, the model fails to effectively utilize the corner regions of the feature map. In contrast, with Elliptical Grid Mapping, most regions of the feature map are efficiently utilized.

For convenience, we have included the core implementation of the Near-Equal-Area Warping method in the supplementary material as \texttt{near\_equal\_area\_warping.py}.

\footnotetext[1]{\url{https://en.wikipedia.org/wiki/Lambert_azimuthal_equal-area_projection}}
\footnotetext[2]{\url{https://github.com/Kuuuube/Circular_Area/blob/main/wiki/mappings/elliptical_grid_mapping.md}}


\section{Details of Hy-Plane (2+2)}
\label{sec:supp_D}

Although the LAEA projection addresses seam artifacts, its implementation still requires "opening" the South Pole, which inherently leaves one remaining polar region. This region is more prone to high-frequency noise and distortion.

While in the Hy-Plane (3+1) formulation we can hide the problematic area by orienting it downward, this approach limits the generality of our representation, particularly for objects or scenes that require rendering from all directions. In such cases, relying on a single hidden region is insufficient, as any arbitrary viewpoint may expose the problematic area and lead to visible artifacts. To make the Hy-Plane representation more universally applicable, we propose Hy-Plane (2+2) to resolve this issue.

The Hy-Plane (2+2) representation combines two planar planes ($P_{XY}$ and $P_{YZ}$) and two spherical planes ($P_a$ and $P_b$). These two spherical planes overlap spatially but are oriented such that their respective South Poles face opposite directions. Specifically, as shown in main paper Fig. 3 (d), the North Pole of $P_a$ is oriented along the negative $z$-axis, while the North Pole of $P_b$ is oriented along the positive $z$-axis; consequently, their South Poles point in opposite directions. By assigning weights to each plane and summing them, the smooth North Polar regions of one spherical plane can effectively cover the problematic South Polar regions of the other, thereby eliminating the distortion-prone areas entirely.







\section{Parameter Count, Speed and VRAM Usage Comparison}
\label{sec:supp_E}

To provide a comprehensive comparison of parameter count, training and inference speed, as well as VRAM usage across different methods and HyPlaneHead configurations, we conducted measurements for each experiment listed in Table 1 of the paper. The results are shown in \cref{tab:param_speed_vram}, where an asterisk (*) denotes that two spherical tri-planes are output simultaneously by a shared branch. 

To provide a comprehensive and fair comparison, we clarify the definitions of the metrics used in our evaluation. Representation Parameters refer to the number of parameters in the feature maps of different tri-plane-like representations (e.g., tri-plane, spherical plane, hy-plane, etc.). Total Learnable Parameters denote the total number of trainable parameters in the entire model architecture, such as EG3D, PanoHead, SphereHead, and HyPlaneHead. Training Speed and Training VRAM Usage are measured on a single V100 GPU with a batch size of 2, representing the average time and memory consumption required to train 1,000 images. Similarly, Inference Speed and Inference VRAM Usage are evaluated under the same hardware setup but with a batch size of 1, by generating 100 images and computing the average time and memory cost per image.

From the statistics in the table, we can observe that different representations vary significantly in terms of feature plane parameter count. The tri-plane uses the fewest parameters (3 × 256 × 256 floating-point values), while the tri-grid with 512×512 resolution uses the most (9 × 512 × 512). Our proposed hy-plane uses 1 × 512 × 512 floating-point values, which is only 1.33 times the number used by the tri-plane. However, the overall difference in total learnable parameters across models is relatively small. The minor variations are mainly due to differences in the final convolutional layer configuration of StyleGAN2, which depends on how each representation is generated.

Notably, the total learnable parameters for entries No. 3 and No. 5 are identical. This is because in experiment No. 3, we did not modify the model code at all, which means only the rendering pipeline was adjusted to use one spherical tri-plane. We reported the actual parameter count to remain consistent with our experimental setup.

In terms of training and inference speed, as well as VRAM usage, there are some differences among the experiments. Overall, however, the additional computational overhead introduced by our innovations is relatively small.

Replacing the tri-plane with hy-plane introduces:
\begin{itemize}
    \item +5.5\% training time,
    \item +3.8\% inference time,
    \item +8.8\% training VRAM,
    \item $-$0.8\% inference VRAM.
\end{itemize}

Adding the evenly split strategy further increases the cost by:
\begin{itemize}
    \item +8.4\% training time,
    \item +6.9\% inference time,
    \item +12.2\% training VRAM,
    \item +16.6\% inference VRAM.
\end{itemize}

When combining with the near-equal-area projection, the overhead becomes:
\begin{itemize}
    \item +0.5\% training time,
    \item +3.5\% inference time,
    \item $-$5\% training VRAM,
    \item $-$5.7\% inference VRAM.
\end{itemize}

\noindent Please note that in some cases, VRAM usage may actually decrease, likely due to memory fragmentation causing inaccuracies in the \texttt{nvidia-smi} measurement.

In summary, while our method does introduce some computational and memory overhead, the increase is relatively modest and justifiable given the significant improvements in disentanglement and reconstruction quality.

\begin{sidewaystable}
\centering
\small
\renewcommand{\arraystretch}{1.6}
\resizebox{\textwidth}{!}{
\begin{tabular}{c|l|l|c|c|c|c|c|c|c}
\hline
\textbf{No.} & \textbf{Representation} & \textbf{Unify-Split} & \textbf{Wrapping} & \textbf{Rep. Params} & \textbf{Total Params} & \textbf{Training Speed (sec/kimg)} & \textbf{Inference Speed (ms/image)} & \textbf{Training VRAM (MiB)} & \textbf{Infer. VRAM (MiB)} \\
\hline
1  & Tri-plane                  & -                & -   & 3$\times$256$\times$256 & 53,174,956 & 180.29 & 42.06 & 6436 & 1103 \\
2  & Tri-plane                  & evenly split     & -   & 1$\times$512$\times$512 & 53,230,222 & 197.89 & 44.17 & 7670 & 1149 \\
3  & Spherical Tri-plane        & -                & -   & 3$\times$256$\times$256 & 54,713,868 & 222.84 & 54.99 & 8048 & 1481 \\
4  & Spherical Tri-plane        & evenly split     & -   & 1$\times$512$\times$512 & 53,234,479 & 198.57 & 48.19 & 7692 & 1097 \\
5  & Dual Spherical Tri-plane   & -                & -   & 6$\times$256$\times$256 & 54,713,868 & 245.79 & 54.87 & 9296 & 1481 \\
6  & Dual Spherical Tri-plane * & -                & -   & 6$\times$256$\times$256 & 53,462,509 & 196.88 & 50.42 & 6810 & 1235 \\
7  & Tri-grid                   & -                & -   & 9$\times$256$\times$256 & 53,741,548 & 198.59 & 45.19 & 6836 & 1245 \\
8  & Tri-plane 512$^2$          & -                & -   & 3$\times$512$\times$512 & 53,423,246 & 181.00 & 47.77 & 6450 & 1333 \\
9  & Spherical Tri-plane 512$^2$& -                & -   & 6$\times$512$\times$512 & 54,962,158 & 182.28 & 59.91 & 6744 & 1103 \\
10 & Tri-grid 512$^2$           & -                & -   & 9$\times$512$\times$512 & 54,002,318 & 266.13 & 58.79 & 8642 & 2463 \\
11 & Hy-plane (3+1)             & -                & -   & 4$\times$256$\times$256 & 53,269,388 & 190.40 & 43.64 & 6966 & 1095 \\
12 & Hy-plane (3+1)             & evenly split     & -   & 1$\times$512$\times$512 & 53,230,222 & 206.54 & 46.27 & 7818 & 1277 \\
13 & Hy-plane (3+1)             & evenly split     & yes & 1$\times$512$\times$512 & 53,230,222 & 207.21 & 47.89 & 7432 & 1205 \\
14 & Hy-plane (3+1)             & area-bias split  & yes & 1$\times$512$\times$512 & 53,230,222 & 226.31 & 49.61 & 7558 & 1321 \\
15 & Hy-plane (2+2)             & evenly split     & yes & 1$\times$512$\times$512 & 53,230,222 & 212.27 & 49.81 & 7780 & 1215 \\
16 & Hy-plane (2+2)             & area-bias split  & yes & 1$\times$512$\times$512 & 53,230,222 & 219.73 & 51.74 & 8012 & 1255 \\
\hline
\end{tabular}
}
\caption{Comparison of different 3D representation configurations in terms of parameter count, training/inference speed, and VRAM usage. }
\label{tab:param_speed_vram}
\end{sidewaystable}



\section{Comparison and Discussion with Related Works}
\label{sec:supp_F}

In this section, we provide a detailed comparison of our hy-plane representation with two closely related recent works: OrthoPlanes \cite{he2023orthoplanes} and SYM3D \cite{yang2024sym3d}, and discussion on compatibility with several tri-plane-related algorithms. 

\subsection{Comparison with OrthoPlanes}
OrthoPlanes enhances the standard tri-plane representation by introducing multiple parallel planes along each Cartesian axis, effectively increasing the capacity of the feature field. This design is conceptually similar to the tri-grid formulation, where additional planar slices are used to capture finer geometric details.

However, this strategy does not address the underlying structural limitations of Cartesian-based projections. Specifically, OrthoPlanes still relies on axis-aligned planar projections in Cartesian Coordination, which inherently suffer from mirroring artifacts. As demonstrated in our main paper (\cref{fig:teaser}), such artifacts manifest as mirroring-face artifacts and unnatural left-right duplication. Our hy-plane representation mitigates this by integrating spherical projection planes that naturally align with the radial symmetry of human heads, thereby disentangling symmetric and asymmetric components more effectively.

Moreover, the increased number of planes in OrthoPlanes leads to higher storage overhead. For downstream applications such as 3D head reconstruction or model initialization (e.g., in Portrait3D \cite{wu2024portrait3d}, AnimPortrait3D \cite{wu2025text}, or ID-Sculpt \cite{hao2025id}), each sample must store 
K times more feature maps (where K is the number of parallel planes per axis), significantly increasing memory and bandwidth requirements. In contrast, our hy-plane uses only four planes (three planar + one spherical) while achieving superior fidelity.

Finally, due to the increased number of planes, it becomes difficult to integrate these approaches with our unify-split strategy. As a result, inter-channel feature penetration remains an issue in these methods.

\subsection{Comparison with SYM3D}
First, our goal differs from that of SYM3D. SYM3D enhances the symmetry of tri-plane representations through symmetric regularization, which is beneficial for generating fully symmetric artificial objects. In contrast, our hy-plane is designed to support a broader range of real-world scenarios where both symmetric and asymmetric structures coexist, such as in full-head portraits. 
Second, regarding feature penetration, SYM3D employs an attention-based scheme (View-wise Spatial Attention) to learn how to alleviate feature penetration across channels, whereas our hy-plane utilizes the unify-split strategy to geometrically and fundamentally prevent feature penetration at its source.

In fact, the correlation across $P_{XY}$, $P_{YZ}$ and $P_{XZ}$ planes discussed in SYM3D essentially corresponds to the inter-channel feature penetration problem we identify in our paper, though they observe it from a different angle. We detect this issue through visual inspection of feature maps (as shown in \cref{fig:teaser}(a,b)), while SYM3D quantifies it using correlation metrics.

Inter-channel feature penetration causes similar values at the same UV positions across different feature maps, resulting in visually similar patterns, exactly what can be seen in our \cref{fig:teaser}(a,b). This phenomenon is also visible in the top-right panel of fig. 7 in SYM3D, where GET3D’s feature maps exhibit repetitive vertical lines at the same spatial locations when zoomed in. Numerically, this leads to high correlations between different feature maps.

The difference lies in terminology: SYM3D does not explicitly refer to this as inter-channel feature penetration, but rather describes it as correlation.
SYM3D attributes this issue to the use of single-view images during training, as opposed to multi-view data. While this is certainly true—more views provide richer geometric cues and reduce ambiguity—we also discuss this point in our paper. From a more fundamental perspective, however, the issue arises due to the structural nature of convolutional networks, where different channels are prone to interfere with each other, especially in the absence of direct supervision. Our solution addresses this root cause directly by modifying the architecture via the unify-split strategy, which effectively eliminates inter-channel feature penetration at its source.

In contrast, SYM3D uses view-wise spatial attention to alleviate the issue. As shown in the bottom-right panel of fig. 7 in SYM3D, this approach does reduce correlation to some extent. However, as illustrated in the middle-bottom panel, strong correlations still exist between certain planes, particularly $P_{YZ}$ and $P_{XZ}$. We have also computed and visualized similarity matrices similar to those in SYM3D's fig. 7, and our results show significantly lower feature correlations across different planes. 

Moreover, in terms of implementation complexity, our unify-split strategy is simpler and more intuitive compared to SYM3D’s view-wise attention mechanism, while achieving a more complete resolution of the problem.


\subsection{Discussion on Compatibility with Tri-plane-related Algorithms}

Dual Encoder \cite{bilecen2024dual} introduces a dual-encoder GAN inversion approach for single-view 3D full-head reconstruction. It uses one encoder for the visible front region and another for the occluded back region, addressing issues such as mirroring artifacts in PanoHead’s W space. In contrast, our HyPlaneHead improves the underlying representation structure to reduce such artifacts and enhance W space quality. As a result, standard inversion methods like PTI—a general-purpose technique for common GANs—already yield better performance than previous approaches. We believe that combining our method with specialized inversion strategies like Dual Encoder, which are specifically tailored for full-head generation, could further improve results. 

Tri2-plane \cite{song2024tri} introduces a cascaded tri-plane representation across multiple scales of facial features, similar to a feature pyramid. This hierarchical design enables the model to generate both global structure and fine-grained details, resulting in richer and more detailed head reconstructions. Although it is still based on the tri-plane formulation and thus remains susceptible to mirroring artifacts, we believe that the multi-scale feature pyramid concept could be beneficially integrated with our hy-plane representation in future work.

Our method is also compatible with recent local editing approaches, such as \cite{bilecen2025reference}, which has been successfully applied to both tri-plane (EG3D) and tri-grid (PanoHead) representations. Given that our hy-plane shares a similar structure, we believe the method should be directly applicable to our representation as well.

Inspired by recent 3D avatar approaches \cite{qian2024gaussianavatars, zhang2025hravatar, qiu2025anigs, kirschstein2025avat3r, chu2024generalizable, qiu2025lhm, kirschstein2024gghead} based on 3D Gaussian Splatting (3DGS) \cite{kerbl20233d}, we believe our hy-plane representation can be effectively integrated with 3DGS to further enhance rendering quality. We plan to explore this promising direction in future work.

Additionally, we believe that replacing the standard tri-plane representation in methods like LRM \cite{hong2023lrm} and InstantMesh \cite{xu2024instantmesh} with our hy-plane has the potential to improve their generation quality. On one hand, our unify-split strategy eliminates inter-plane feature penetration, allowing each plane to express its features more clearly and effectively. On the other hand, by incorporating a spherical plane, we enhance the model’s ability to represent asymmetric regions, such as facial details and hair on the back of the head. Therefore, we expect that integrating hy-plane into these models would lead to more accurate and artifact-free 3D reconstructions.

\section{More Qualitative Comparison}
\label{sec:supp_G}

\begin{figure}[h!]
    \centering
    \includegraphics[width=1\linewidth]{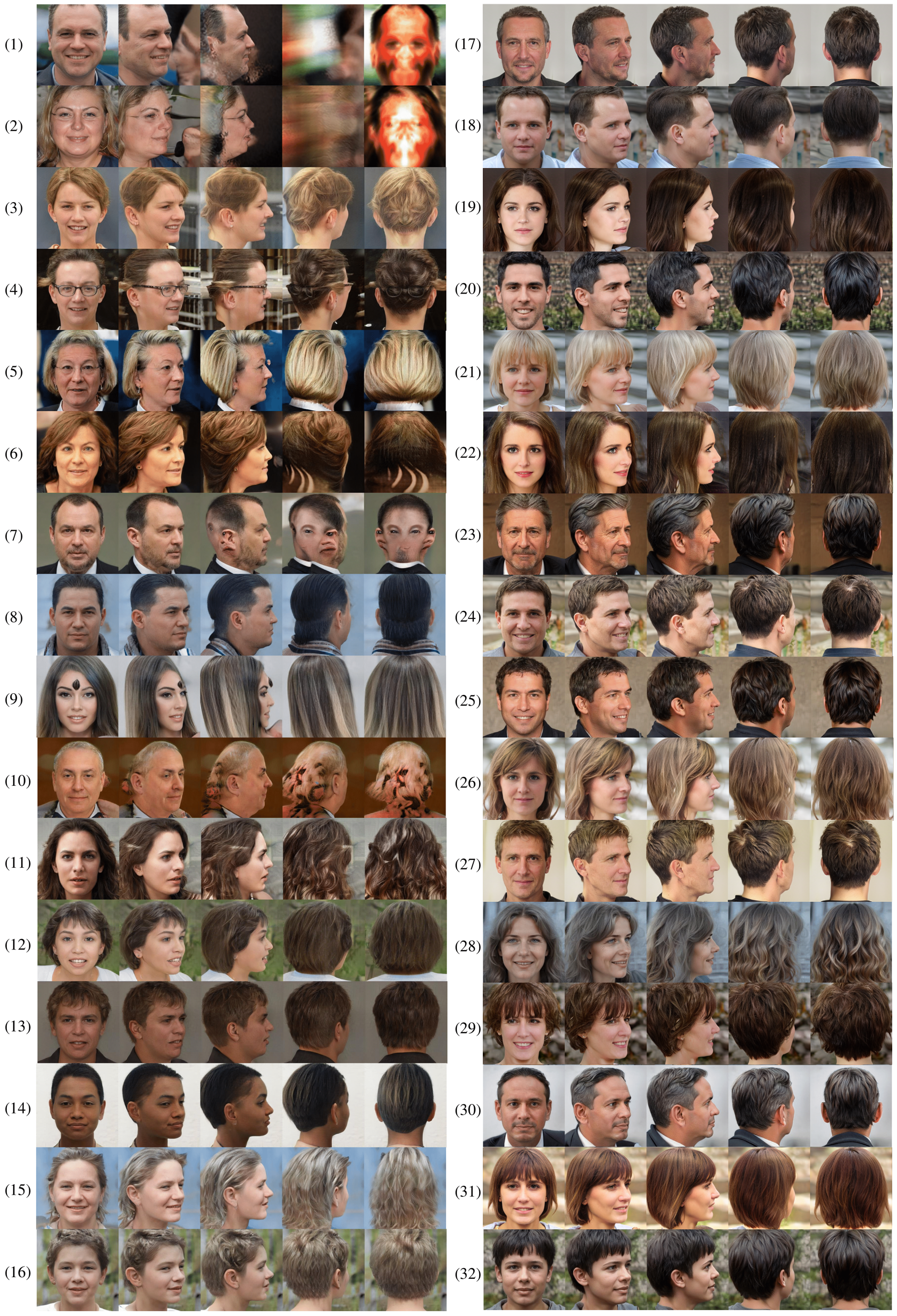}
    \caption{Additional qualitative results.}
    \label{fig:more_qualitative}
\end{figure}

Figure~\ref{fig:more_qualitative} shows additional qualitative comparisons between our method and existing approaches on a larger set of examples. On the left side, samples (1–2) are generated by the official EG3D model, which is trained exclusively on the FFHQ dataset that lacks large-pose and back-view head images. As a result, these samples exhibit severe mirroring artifacts in their back views. Samples (3–4) use the tri-plane representation trained within our pipeline and data; while hair appears in the rear region, the front-facing facial attributes still dominate, indicating incomplete adaptation to non-frontal views. Samples (5–7) come from the official PanoHead model, which can produce detailed facial and hair textures but suffers from strong left-right symmetry and visible artifacts in the back view. Similarly, samples (8–10), using the tri-grid representation trained with our setup, also exhibit comparable symmetry and artifact issues. These problems—mirroring artifacts and severe left-right symmetry—are primarily caused by the Cartesian projection used in both tri-plane and tri-grid representations. Samples (11–13) are from the official SphereHead model, which addresses the mirroring issue through its spherical tri-plane design but results in more blurred outputs with less detailed facial and hair textures. The same trend is observed in samples (14–16), where a spherical tri-plane model is trained using our pipeline and data. In contrast, on the right side, samples (17–32) are generated by our HyPlaneHead model. By leveraging the novel hy-plane representation, our method not only eliminates mirroring artifacts but also achieves high-quality, consistent rendering from arbitrary view angles.

\section{Limitations}
\label{sec:supp_H}

Despite the promising results of our method, there are still several limitations that warrant further investigation. First, similar to previous works such as EG3D, PanoHead, and SphereHead, our model exhibits minor visual flickering or instability in fine details when rendering from gradually changing viewpoints. We believe this is partly due to the current GAN backbone's limited capacity for high-fidelity view-consistent generation, and we plan to address this by adopting a more powerful generator architecture in future work.

Second, our method, like existing approaches, struggles with generating highly complex hairstyles such as ponytails, braids, or other structured hair arrangements. This limitation likely stems from insufficient training data covering such styles. We regard this as an important direction for future research and intend to expand our training dataset to include a broader variety of hairstyles and appearances.

\section{Code of Ethics}
\label{sec:supp_I}

Our work presents a method for learning generalizable 3D full-head modeling from monocular images, which has potential applications in virtual avatars, digital content creation, and immersive experiences. However, such technology also raises ethical concerns, particularly regarding privacy and the potential for misuse, such as identity deception or unauthorized generation of realistic 3D head models. We are aware of these risks and emphasize the importance of responsible deployment, transparency, and user consent in any real-world application of this technology.